\documentclass[journal]{IEEEtran}

\usepackage{cite}
\usepackage{graphicx}
\usepackage{amsmath}
\usepackage{algorithm}
\usepackage{tikz}
\usepackage{amssymb}
\usepackage{physics}
\usepackage{lipsum}
\usepackage{mathtools}
\usepackage{siunitx}
\usepackage{wasysym}
\usepackage{array}

\DeclareRobustCommand{\ppp}[0]{%
\begin{tikzpicture}[line width=0.3pt, scale=1.2]%
\draw (0ex,0ex) -- (1.5ex,0ex);
\draw (1.5ex,0ex) -- (0.75ex,1.5ex);
\draw (0.75ex,1.5ex) -- (0ex,0ex);
\end{tikzpicture}}

\DeclareRobustCommand{\pppp}[0]{%
\begin{tikzpicture}[line width=0.3pt, scale=1.2]%
\draw (0ex,0ex) -- (1.5ex,0ex);
\draw (1.5ex,0ex) -- (0.75ex,1.5ex);
\draw (0.75ex,1.5ex) -- (0ex,0ex);
\draw (0.75ex,0.65ex) -- (0ex,0ex);
\draw (0.75ex,0.65ex) -- (1.5ex,0ex);
\draw (0.75ex,0.65ex) -- (0.75ex,1.5ex);
\end{tikzpicture}}

\usepackage{tabularx}
\usepackage{multirow}
\usepackage{array}
\usepackage{hyperref}
\newcolumntype{L}[1]{>{\raggedright\let\newline\\\arraybackslash\hspace{0pt}}m{#1}}
\newcolumntype{C}[1]{>{\centering\let\newline\\\arraybackslash\hspace{0pt}}m{#1}}
\newcolumntype{R}[1]{>{\raggedleft\let\newline\\\arraybackslash\hspace{0pt}}m{#1}}
\newcolumntype{Y}{>{\centering\arraybackslash}X}

\newcommand\myeq{\stackrel{\mathclap{\scriptsize\mbox{def}}}{=}}

\begin{document}

\title{Malleable Robots: Reconfigurable Robotic Arms with Continuum Links of Variable Stiffness}

\author{Angus~B.~Clark,~\IEEEmembership{Student~Member,~IEEE,}
        and~Nicolas~Rojas,~\IEEEmembership{Member,~IEEE}
\thanks{This paper includes a supplementary video available at \url{https://youtu.be/Yqvjom6As6U}.}
\thanks{Some parts of this paper were presented at the IEEE International Conference on Robotics and Automation, Montreal, Canada, May 20-May 24, 2019 and Paris, France, May 31-August 31, 2020.}
\thanks{Angus B. Clark and Nicolas Rojas are with the REDS Lab, Dyson School of Design Engineering, Imperial College London, 25 Exhibition Road, London, SW7 2DB, UK
{\tt\small (a.clark17, n.rojas)@imperial.ac.uk}}
}




\maketitle

\begin{abstract}
Through the implementation of reconfigurability to achieve flexibility and adaptation to tasks by morphology changes rather than by increasing the number of joints, \emph{malleable robots} present advantages over traditional serial robot arms in regards to reduced weight, size, and cost. While limited in degrees of freedom (DOF), malleable robots still provide versatility across operations typically served by systems using higher DOF than required by the tasks. In this paper, we present the creation of a 2-DOF malleable robot, detailing the design of joints and malleable link, along with its modelling through forward and inverse kinematics, and a reconfiguration methodology that informs morphology changes based on end effector location---determining how the user should reshape the robot to enable a task previously unattainable. The recalibration and motion planning for making robot motion possible after reconfiguration are also discussed, and thorough experiments with the prototype to evaluate accuracy and reliability of the system are presented. Results validate the approach and pave the way for further research in the area. 
\end{abstract}


%
\IEEEpeerreviewmaketitle

\section{Introduction}
\IEEEPARstart{R}{econfigurable} robot systems provide several key potential advantages over traditional robots, including increased task versatility by adapting to better suit tasks, and reduced robot cost due to a smaller total number of modules, such as links and joints. As such, there has been significant research into the development of reconfigurable robots, with the most popular approach utilising modularity as the method of reconfiguration, as this allows for the interchangeability of parts, leading to self-repair \cite{yim2007modular, seo2019modular}. The reconfigurability feature has specifically been of interest in unstructured and unpredictable environments, characterised by changing operating contexts, which take the most advantage from robots that can adapt their shape and operating mode \cite{valente2016reconfigurable}. Furthermore, environments that have limited access provide a strong argument for reconfigurable robots that can fold or disassemble to squeeze through small holes, such as for the inside of aircraft wings \cite{roy2005design}.

Regarding the structure of robots for reconfiguration, the ModMan serial manipulator for instance utilises a modular system with a variety of modules for links and joints \cite{yun2020modman}. A similar design is also presented by Xu et al \cite{xu2021wireless} and Strasser et al \cite{strasser2008distributed}, whereas robots like M-TRAN \cite{murata2002m}, PolyBot \cite{yim2000polybot}, and ATRON \cite{brandt2007atron} use a single modular component, reducing the cost of production and increasing the range of abilities of the robot (such as locomotion), albeit with a decrease in the performance for a specific task compared to a specialised robot. While the majority of reconfigurable robots are modular, reconfiguration can also be achieved by locking aspects of the robot. Examples include directly locking revolute joints to reduce the DOF of the robot \cite{lu2016design}, and locking passive cylindrical joints carefully positioned to directly vary the Denavit-Hartenberg (DH) parameters of a serial arm \cite{aghili2009reconfigurable, ananiev2002approach}. Reconfigurable robots are also not just limited to serial manipulators, parallel reconfigurable robots have also been demonstrated, also using a similar joint locking method to achieve alternative configurations \cite{plitea2013structural}. They have also been demonstrated using a modular structure with variable-dimension rigid links \cite{velasco1998automated}. In the case of the Tricept-IV robot, a hybrid reconfigurable robot is achieved through the combination of both serial and parallel mechanisms, which exploits the advantages of both---namely, the stiffness and accuracy of the parallel mechanism and the workspace size of the serial mechanism \cite{sun2010workspace}.

\begin{figure}[t!]
    \centering
    \includegraphics[width=0.95\columnwidth]{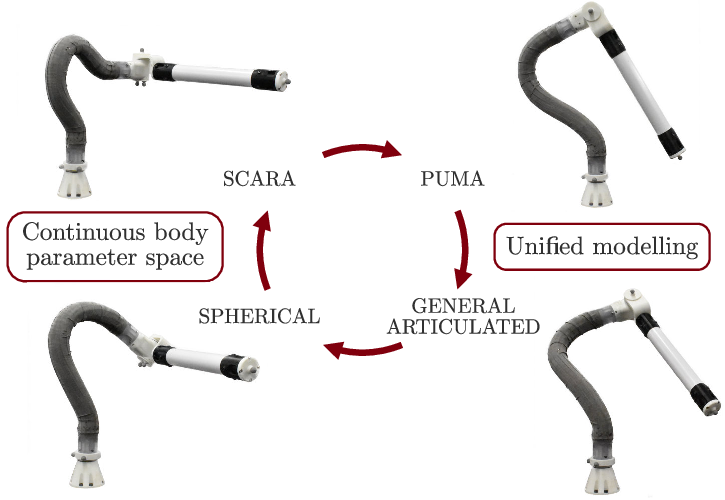}
    \caption{A 2-DOF malleable robot visual description of capability, demonstrating the possible topology types of the malleable robot.}
    \label{malleablerobotdiagram}
\end{figure}

An alternative approach for the application of reconfigurable robot manipulators can be found in the industrial field of serial manipulators. In the ideal case, a manipulator would be designed with the exact number and configuration of joints necessary for its expected set of tasks \cite{kereluk2017task}. This is known as task-based optimisation, and requires information to be known about the robot structure \cite{brandstotter2018task}, collections of working points \cite{yang2000task}, or end effector regions \cite{kereluk2017task}. However, knowledge of all tasks a robot might encounter in its lifetime can be difficult to determine. Instead, serial manipulators with a higher degree of freedom (DOF) are typically selected, ensuring dexterity across tasks at the expense of an increased robot cost and footprint. 

Malleable robots, which are reduced DOF serial robot arms with changeable geometry, provide a solution to the task versatility problem through variable relative positioning of the revolute joints \cite{clark2020design}. An example of the reconfiguration capability of a 2-DOF malleable robot can be seen in Fig.~\ref{malleablerobotdiagram}. A method for achieving this capability is through continuously bending variable stiffness links, which have been demonstrated using vacuum-activated layer jamming \cite{clark2020design, zhou2020discrete, clark2019stiffness} and granular jamming \cite{clark2019assessing}, and LMPAs (low melting point alloys) \cite{brandstotter2018task}. It is worth highlighting that many other mechanisms exist for achieving variable stiffness, as demonstrated by Blanc et al. \cite{blanc2017flexible} in the medical robotics field, however the larger size of robotic links needed for other applications limits the possible technologies usable \cite{clark2019assessing}. 

Malleable robots follow a similar process of reconfiguration to modular reconfigurable robots, in that their reconfiguration alters the relative positioning of the active joints of the robot. However, unlike modular reconfigurable robots which achieve this by disassembling and reassembling (in an alternative configuration) the modules of the robot, malleable robots achieve this by transitioning between reconfigurable and rigid modes. Once in a reconfigurable mode, the robot can either reconfigure itself using additional drive systems (intrinsic malleable robots), or it can be manually reconfigured by an external system, such as a user (extrinsic malleable robots) as discussed in \cite{clark2020design}, where the design of such robots was firstly explored. Open questions regarding the modelling and optimisation of the topology of malleable robots are particularly addressed herein, presenting alongside improvements in their design.

Multiple methods for solving the reconfiguration planning problem, that is, determining the ideal configuration of a robot based on a course of actions \cite{seo2019modular}, have been presented for modular reconfigurable robots. One method developed is the optimisation to minimise joint torques based on point masses of the modules \cite{dograoptimal}. It has been shown that the control of modular reconfigurable robots based on joint torque sensing is robust in that modules can be added or removed without the need to adjust control parameters \cite{liu2008distributed, xu2021wireless}. Due to the complex nature of determining an ideal reconfiguration, one solution presented is the use of a digital twin, a set of virtual information constructs that fully describe the reconfigurable machine, which can be used to simulate reconfigurations to determine an optimum configuration \cite{huang2020building}. 

For a modular reconfigurable robot composed of rigid links and joints, a genetic algorithm has been proposed based on the Jacobian matrix that finds a near-optimal solution for link lengths, having determined the DOF of the robot based on the task specification \cite{chung1997task}. Using simple cubic and tri-prism modules, a similar approach whereby reconfiguration is determined based on a single tool position for each configuration has also been presented \cite{ding2013fundamental}. A genetic algorithm is additionally presented for determining the passive/active joint configuration of a modular reconfigurable robot \cite{zhang2018underactuated}. Herrero et al. present a grasp point optimisation method that determines the best configuration of a 6 DOF parallel manipulator, determined by the size and regularity of the resulting workspace, that considers multiple end-effector locations \cite{herrero2015enhancing}. A spherical area is generated from the desired grasp points, which is then used to find an optimal workspace based on the geometric and actuator restrictions of the robot.

Modelling the resulting new geometric topology of reconfigurable robot manipulators, i.e. determining the forward kinematics (FK) and inverse kinematics (IK), is additionally a challenge. Determining the resulting workspace of such robots can then either be performed using the calculated FK (for a discrete solution), or it can present another challenge to overcome for a continuous solution. For example, Chen et al. presents a recursive Newton-Euler algorithm that constructs the equations of motion of a tree-structured modular reconfigurable robot described using simplified vertices and edges \cite{chen1998automatic}. For developed reconfigurable robot manipulators, direct internal measurement of their reconfiguration can be challenging. The use of external motion tracking cameras is a popular mechanism for aiding modelling of the robot, and has been utilised in autonomously identifying the kinematic model (that is, determining the robot kinematic structure, followed by an estimation of the forward kinematics) of a robot where no prior knowledge of the robot geometry is assumed \cite{dalla2020autonomous}.

\begin{figure}[t!]
    \centering
    \includegraphics[width=\columnwidth]{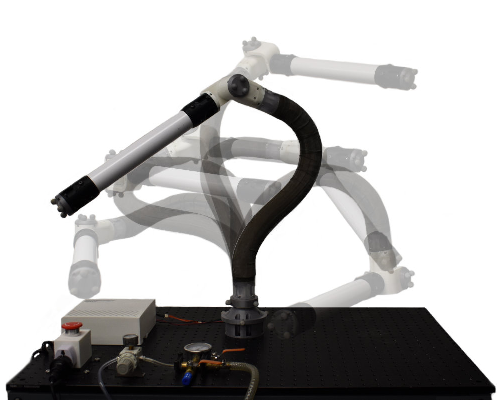}
    \caption{The developed two-degree-of-freedom (DOF) malleable robot arm, showing various topology configurations it can achieve. A PUMA-like configuration is shown in foreground.}
    \label{main}
\end{figure}

\begin{figure*}[t!]
    \centering
    \includegraphics[width=\textwidth]{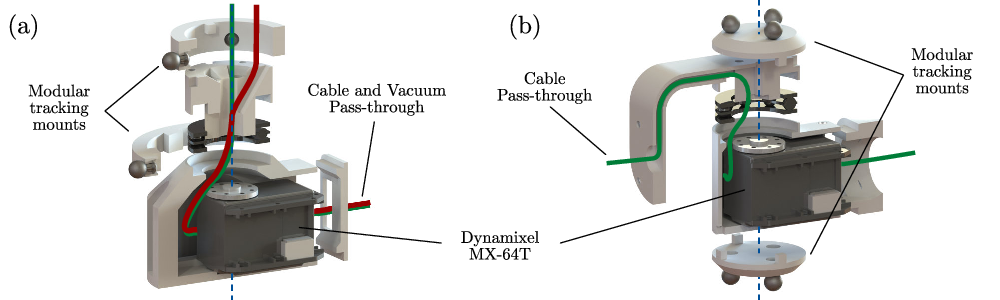}
    \caption{Exploded CAD of the two revolute joints, \textbf{(a)} joint 1 and \textbf{(b)} joint 2, of the malleable robot, showing highlighted cable and vacuum pass-through.}
    \label{joints}
\end{figure*}

\begin{figure*}[t!]
    \centering
    \includegraphics[width=\textwidth]{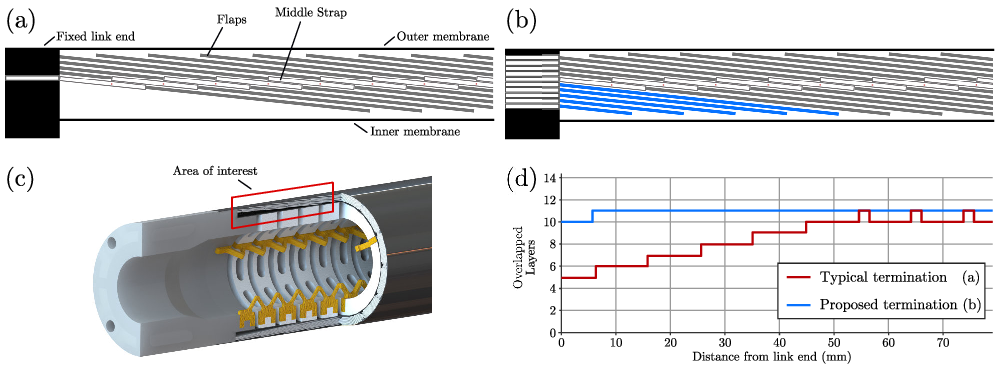}
    \caption{Comparison of the layer jamming termination methods: \textbf{(a)} A typical termination performed by trimming the layers to the desired length, \textbf{(b)} the proposed layer termination method with added passive layers, \textbf{(c)} cross-sectional view of the malleable link showing the layer jamming area of interest, and \textbf{(d)} a plot of overlapped layers against distance from link end, comparing the typical termination \textbf{(red)} and the proposed termination with added passive layers \textbf{(blue)}. In the typical design a decrease in layers towards the link end is shown, as well as the misalignment of layers resulting in the oscillation between 10--11 overlapped layers far from the link end.}
    \label{layertermination}
\end{figure*}

In this paper, we introduce a topology optimisation method for malleable robots based on distance geometry, which is capable of determining feasible robot geometric topologies (configurations of the robot joints defined by interpoint-distances) based on a desired end effector point. The developed robot is shown in Fig.~\ref{main}. A single vector defining the end effector location and orientation is required for the feasible topologies, and once determined the malleable robot can be extrinsically reshaped using real-time motion tracking feedback to aid alignment to the desired topology. 

This paper is an evolved paper from a previous article that presented a preliminary design of a 2-DOF malleable robot, along with its workspace definition \cite{clark2020design}. This paper improves and refines this design, and introduces the computation of the forward and inverse kinematics as well as the topology reconfiguration of the robot.

The rest of this paper is organised as follows. In section II, the improved design of a malleable robot is explored and the developed malleable robot is described. In section III and IV, the distance-based parameters and main set of formulae defining the malleable robot workspace are identified, and then the symbolic equation of the workspace surface traced by the end effector of a 2-DOF malleable robot is obtained, along with a presentation of its workspace categories. In section V, we compute the forward and inverse kinematics of the malleable robot based on trilateration. In section VI we describe the developed topology optimisation methodology and subsequent robot joint calibration. In sections VII and VIII, we evaluate and comment on the performance of the developed kinematics and topology optimisation. Finally, we conclude in section IX.

\section{Malleable Robot Design}
A 2-DOF malleable robot, formed from two revolute joints, a malleable link, and a rigid link, was developed following our previous work \cite{clark2020design}. Joint 1, positioned at the base of the robot, provides rotation in the z-axis. Joint 2 was positioned at the end of the malleable link, providing rotation in the axis perpendicular to the termination end. Both joints were constructed from a Dynamixel MX-64 servo motor, with a 3D printed ABS housing, and a thrust ball bearing (size 51106) providing force distribution of the motor torque to the output side of the joint. The malleable link is a variable stiffness link that uses Mylar-based layer jamming to transition between rigid and flexible modes, with a maximum length of 700~mm and a minimum length of 550~mm. The rigid link attached to the secondary joint has a length of 370 mm (actual distance of 450 mm between joint axis and end effector). The link was composed of a 42 mm~$\diameter$ polypropylene tube, and was attached to the robot using 3D printed ABS link ends similar to those used on the malleable link. The dimensions of the distal link were selected to be like that of the malleable link, with the shorter length chosen to prevent frequent collisions with the floor plane. Details of the joints and malleable link are discussed next. 


\begin{figure}[t!]
    \centering
    \includegraphics[width=0.9\columnwidth]{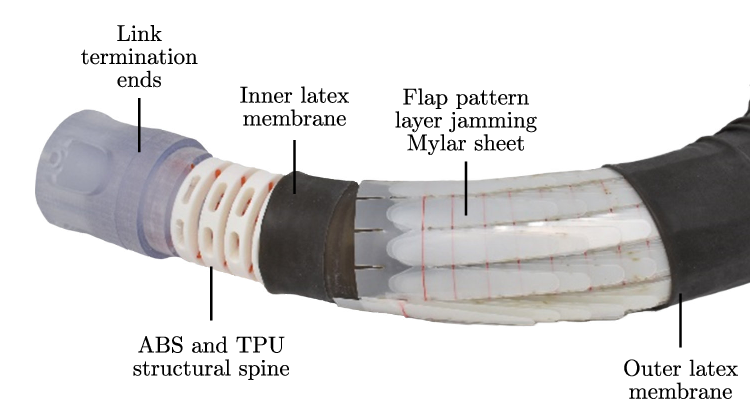}
    \caption{Partially constructed malleable link highlighting the design of the variable stiffness components.}
    \label{malleablelink}
\end{figure}

\subsection{Joints}
In order to provide robust motion tracking of the manipulator geometry, modular tracking mounts were added to both revolute joints along the axis of rotation. On these, mounts with 3 reflective markers were placed equidistant around the revolute axes, allowing for the creation of a rigid body with a central point located on the axis for each mount. For joint 1, this required markers to surround the joint, due to the positioning limitations imposed by the fixation to a base and malleable link. For joint 2, this was straightforward due to the free space either side of the aligned input-output joint. The design of the malleable robot also necessitated a 'clean' implementation, with any pass-through elements, such as wires for controlling an end effector, passing through the axis of the joints rather than as an external system routed along the robot topology. This was done to prevent any change in length of such systems experienced under certain configurations, which would result in limitations being placed on the topology reconfiguration of the robot. A similar issue regarding tendon lengths in a reconfigurable gripper was addressed by Lu et al. \cite{luruth}. For joint 1, this also required partially enlarging and reinforcing the design to accommodate the vacuum tube used to control the stiffness of the malleable link. The modular tracking mounts, along with the cable and vacuum pass-through design, can be seen in Fig.~\ref{joints}.

\subsection{Malleable Link}
\begin{figure}[t]
    \centering
    \includegraphics[width=0.95\columnwidth]{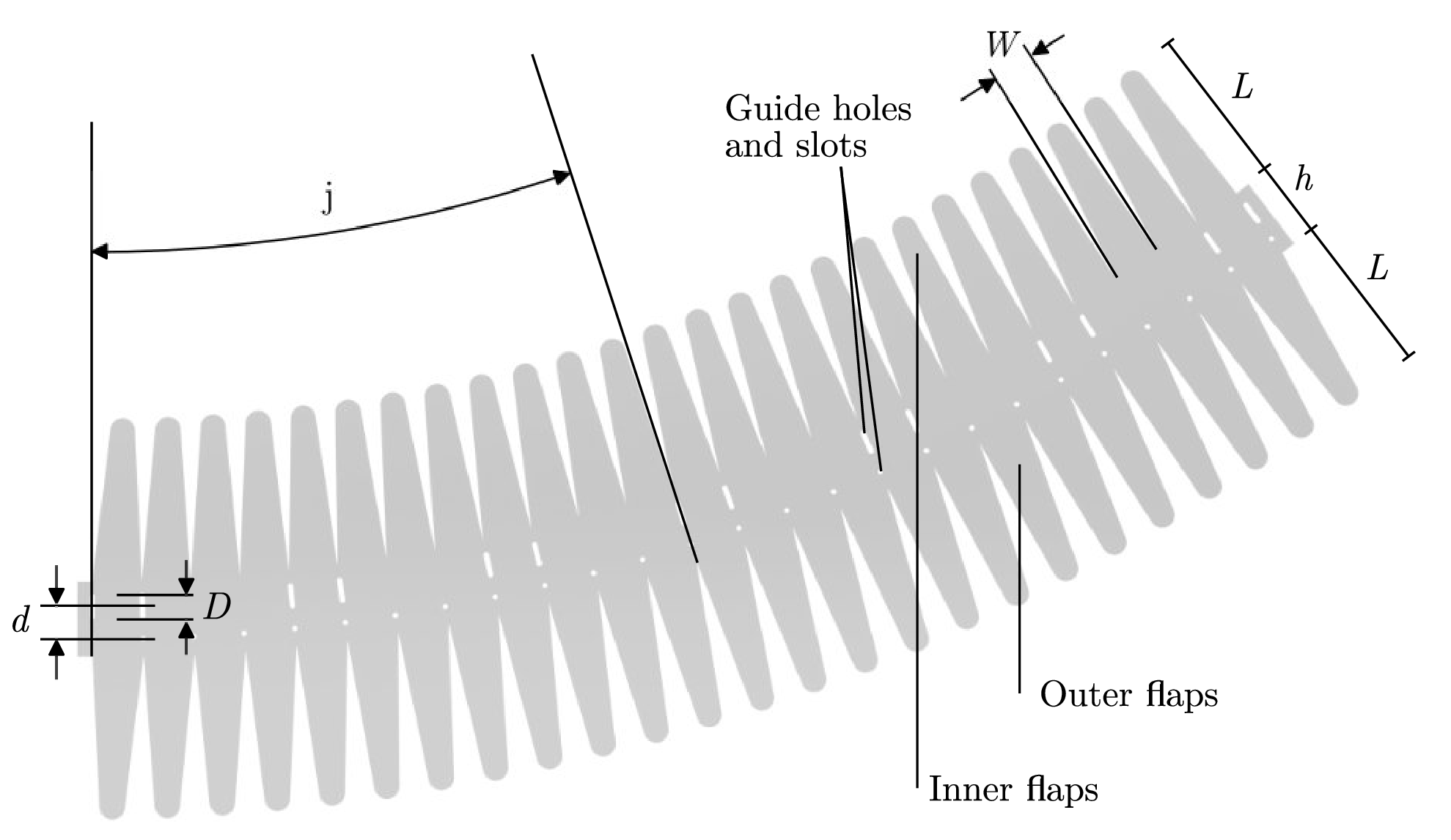}
    \caption{Double-sided flap pattern specifications for layer jamming sheath with guide holes and slots.}
    \label{LayerDesign}
\end{figure}

For the malleable link, layer jamming using conically wrapped overlapping sheets of Mylar\textsuperscript{\textregistered} (0.18~mm) (polyethylene terephthalate), first presented as a medical manipulator by Kim et al. \cite{kim2012}, and later enlarged for a malleable link in our previous research \cite{clark2019assessing} was used. The laser cut flap pattern, detailed in Fig.~\ref{LayerDesign}, contained 12 flaps spanning the circumference of the link, with a minimum of 10 overlapping layers always in contact. Flap parameters used were flap length $L=45mm$, flap width $W=13mm$, mid length $h=16mm$, guide hole distance $d=9.5mm$, and inclination angle $\varphi=12.75^\circ$. The flap pattern was then wrapped conically and contained within two cylinder membranes of latex sheet (0.25 mm), and sealed with link termination ends 3D printed from Vero Clear on a Stratasys Objet 500, which also provided mounting points for an internal structural spine to prevent excessive deformation under extreme bending of the link, as well as mounting points to attach the other components of the robot. The internal spine, developed in our previous work \cite{clark2019stiffness}, was 3D printed from Acrylonitrile Butadiene Styrene (ABS), with flexible couplings connecting the spinal segments printed from Ninjaflex material. By changing the pressure inside the sealed latex membranes using a vacuum pump (BACOENG 220V/50Hz BA-1 Standard), the Mylar layers compress together, and the cumulative friction causes a significant increase in rigidity, proportional to the negative pressure applied. The components of the implemented malleable link can be seen in Fig.~\ref{malleablelink}. The 700~mm length of the malleable link was selected based on the minimum bending radius of the link, such that an equivalent bending performance to a 3-segment continuum robot was achieved, where each segment is capable of 90$^\circ$ bending. As the malleable link is a single structure, a bend capability of 270$^\circ$ was implemented.

While layer jamming provides a high level of stiffness when a vacuum is applied between the encapsulating membranes, the point at which the layer jamming fixes to the rigid joints of the robot (which we call 'layer termination') presents a significant reduction in that stiffness. In typical layer jamming, due to the conical design, a fixture point is not obvious. Therefore, the typical solution is to cut the tubular structure to form a flat face, which can then be adhered to. This is shown in Fig.~\ref{layertermination}(a). As the flap pattern is fixed at the middle strap, this cut results in the removal of $\sim$half of the overlapped layers at the termination point, which is reached by a progressive decrease in overlapped layers from $\sim$45~mm from the termination point. This decrease is shown in Fig.~\ref{layertermination}(d). To correct this, we introduce passive layers as a separate system to the layer jamming, which fill the position of the removed layers. These are fixed to the link end, at the position in red shown in Fig.~\ref{layertermination}(c), and do not extend/compress with the layer jamming--however they mitigate the described loss of stiffness and allow bending when stiffness is deactivated. The added passive layers are shown in Fig.~\ref{layertermination}(b) in blue, and the resulting increase in overlapped layers is shown in Fig.~\ref{layertermination}(d). The transition as an outer flap finishes and an inner flap begins also results in a variation of the overlapped layers, shown in Fig.~\ref{layertermination}(d) at distances greater than $\sim$45~mm. This was additionally corrected by slightly increasing the length of the flaps by 2~mm.

\section{Overview of Distance Geometry}
Distance geometry, first defined by Leonard Blumenthal in first half of the 20th century \cite{blumenthal1953theory}, is a relatively new branch of mathematics that focuses on the study of geometries through the use of their metrics -- i.e. their distances. It avoids the need to define arbitrary reference frames, and the combination of representations of rotations and translations in the generation of equations \cite{rojas-thesis}, as it occurs in a kinematic modelling using Denavit-Hartenburg (D-H) parameters. In this paper, distance geometry is used as an alternative to other methods for computing the kinematics and topology reconfiguration, particularly D-H parameters, as it avoids further hardware and software complexities. For example, it may be challenging to directly obtain the relative joint positions and angles using internal sensors in the malleable robot for D-H modeling, which is an ongoing area of research in continuum robots for instance \cite{wang2017soft}. In contrast, distance geometry only requires the localisation of points distributed along the robot, which can simply be achieved with external motion tracking. This is especially key for the computation of workspaces of malleable robots as D-H parameterization \cite{kung2005development, li2011design}, cannot be directly employed since both link dimensions and the relative orientation of the joints can change. Suitable approaches to perform the workspace analysis are screw theory \cite{coppola20146, xie2015design} or distance geometry \cite{rojas-thesis} since the parameterisation does not depend on relative angles and distances between joint reference frames. We make use of a distance-geometry-based method herein as the technique has been shown to simplify the computation of the workspace equation of complex mechanisms \cite{rojas-coupler, rojas-peaucellier}.

\subsection{Notation}
In what follows, we denote a point in $\mathbb{E}^3$ as $P_i$, $\mathbf{p}_{i,j}=\overrightarrow{\rm P_iP_j}$ denotes the vector from $P_i$ to $P_j$, $\mathbf{p}_{i,j,k}=\mathbf{p}_{i,j}\times\mathbf{p}_{i,k}$ denotes the cross product between vectors $\mathbf{p}_{i,j}$ and $\mathbf{p}_{i,k}$, and $s_{i,j}=\norm{\mathbf{p}_{i,j}}^2=d_{i,j}^2$ denotes the squared distance between $P_i$ and $P_j$, with vector coordinates arranged as column vectors. The vectors $\mathbf{p}_{i,j}$, $\mathbf{p}_{i,k}$, and $\mathbf{p}_{i,j,k}$ in general represent a non-orthogonal reference frame that is denoted by the column vector of nine components $\mathbf{q}_{i,j,k}=(\mathbf{p}_{i,j}^T, \mathbf{p}_{i,k}^T, \mathbf{p}_{i,j,k}^T)^T$.

\begin{figure}[t!]
    \centering
    \includegraphics[width=0.8\columnwidth]{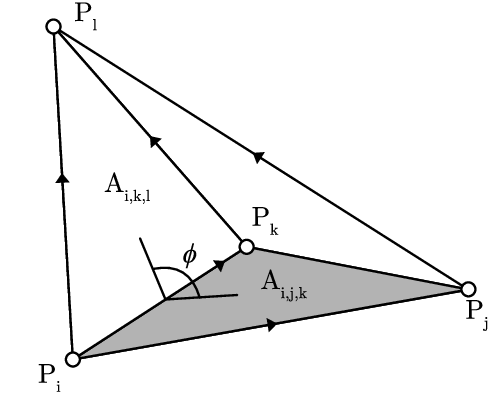}
    \caption{A dihedral angle $\phi$ of the tetrahedron $\pppp_{i,j,k,l}$ defined by the two triangles $\ppp_{i,j,k}$ and $\ppp_{i,k,l}$. Base vectors $\mathbf{p}_{i,j}$ and $\mathbf{p}_{i,k}$ and output vectors $\mathbf{p}_{i,l}$, $\mathbf{p}_{j,l}$, and $\mathbf{p}_{k,l}$ are shown. In this case, $V_{i,j,k,l}>0$.}
    \label{planeangle}
\end{figure}

The tetrahedron defined by points $P_i$, $P_j$, $P_k$, and $P_l$ is denoted as $\pppp_{i,j,k,l}$, with its \textit{origin} located at $P_i$, its \textit{base} given by the triangle $\ppp_{i,j,k}$ with area $A_{i,j,k}$, \textit{base vectors} $\mathbf{p}_{i,j}$ and $\mathbf{p}_{i,k}$, and \textit{output vectors} $\mathbf{p}_{i,l}$, $\mathbf{p}_{j,l}$, and $\mathbf{p}_{k,l}$. This notation is shown in Fig.~\ref{planeangle}~\cite{rojas2017distance, rojas2018forward}.

\subsection{Cayley-Menger Determinants}
The \textit{Cayley-Menger bideterminant} of two sequences of \textit{n} points, $[P_{i,1},...,P_{i,n}]$ and $[P_{j,1},...,P_{j,n}]$, is defined as \cite{rojas-thesis}

\small
\begin{equation*}
    D(i_1,...,i_n;j_1,...,j_n)=2\left(\frac{-1}{2}\right)^n\left| \begin{array}{cccc}
                0       &   1               & \cdots    & 1             \\
                1       &   s_{i_1,j_1}     & \cdots    & s_{i_1,j_n}   \\
                \vdots  &   \vdots          & \ddots    & \vdots        \\
                1       &   s_{i_n,j_1}     & \cdots    & s_{i_n,j_n}   \\
                \end{array} \right|.
\end{equation*}
\normalsize

When the two point sequences are the same, $D(i_1,...,i_n;i_1,...,i_n)$, this is abbreviated as $D(i_1,...,i_n)$, known as the \textit{Cayley-Menger determinant}. For example, for the 5 points $D(P_1,\ldots,P_5)$ this is

\small
\begin{equation}
    \begin{split}
    D(1,2,3,4,5)&=-\frac{1}{16}\left| \begin{array}{cccccc}
                0   &   1         & 1       & 1       & 1       & 1       \\
                1   &   0         & s_{1,2} & s_{1,3} & s_{1,4} & s_{1,5} \\
                1   &   s_{1,2}   & 0       & s_{2,3} & s_{2,4} & s_{2,5} \\
                1   &   s_{1,3}   & s_{2,3} & 0       & s_{3,4} & s_{3,5} \\
                1   &   s_{1,4}   & s_{2,4} & s_{3,4} & 0       & s_{4,5} \\
                1   &   s_{1,5}   & s_{2,5} & s_{3,5} & s_{4,5} & 0       \\
                \end{array} \right|.
    \end{split}
    \label{eq-CMD}
\end{equation}
\normalsize

For the general point sequence $P_{1}$, $P_{2}$,$\ldots$,$P_{n}$, the Cayley-Menger determinant gives $(n-1)!^2$ times the squared hypervolume of the simplex spanned by the points in $\mathbb{E}^{n-1}$ \cite{Menger}. Hence, $D(1,2,3,4,5) = 0$ in $\mathbb{E}^3$. Similarly, for $n=3$, we have \cite{thomas2005revisiting}, 

\begin{equation}
    D(i,j,k)=4A_{i,j,k}^2=\norm{(P_j-P_i)\times(P_k-P_i)}^2,
    \label{eq1}
\end{equation}

which is the Heron's formula relating the area $A_{i,j,k}$ of triangle $\ppp_{i,j,k}$. This can also be expressed purely in interpoint distances as:
\begin{multline}
    A_{i,j,k}=(p_{i,j,k}(p_{i,j,k}-\norm{\textbf{p}_{i,j}})\\(p_{i,j,k}-\norm{\textbf{p}_{i,k}})(p_{i,j,k}-\norm{\textbf{p}_{j,k}}))^\frac{1}{2},
    \label{eq-tri-area}
\end{multline}
where $p_{i,j,k}$ is half the perimeter of the triangle $\ppp_{i,j,k}$ defined as
\begin{equation}
    p_{i,j,k}=\frac{1}{2}\left(\norm{\textbf{p}_{i,j}}+\norm{\textbf{p}_{i,k}}+\norm{\textbf{p}_{j,k}}\right).
\end{equation}

For $n=4$ we obtain the orientated volume $V_{i,j,k,l}$ of the tetrahedron $\pppp_{i,j,k,l}$ as

\begin{equation}
    D(i,j,k,l)=36V_{i,j,k,l}^2.
    \label{eq2}
\end{equation}
It is defined as positive if $\abs{\mathbf{p}_{i,j},\mathbf{p}_{i,k},\mathbf{p}_{i,l}}>0$, and negative otherwise \cite{rojas2017distance}. For \textit{Cayley-Menger bideterminants}, for $n=3$ we have 

\begin{align}
    D(i,j,k;i,k,l)&=4A_{i,j,k}A_{i,k,l}\cdot\cos{(\phi_{i,j,k,l})} \nonumber \\
    &=((P_i-P_k)\times(P_j-P_k))\cdot \nonumber\\
    &\,\,\,\,\,\,\,\,((P_i-P_l)\times(P_k-P_l)),
    \label{eq3}
\end{align}
where $\phi_{i,j,k,l}$ is the dihedral angle between the two planes defined by the triangles $\ppp_{i,j,k}$ and $\ppp_{i,k,l}$. This can be seen in Fig.~\ref{planeangle}.

\subsection{Trilateration}
Trilateration is a method for computing the location of an unknown point using known distances of the point from 3 different known sites. For example, given a tetrahedron $\pppp_{i,j,k,l}$ (Fig.~\ref{planeangle}), we can compute the output vector $\textbf{p}_{i,l}$ as \cite{rojas2017distance}

\begin{equation}
    \textbf{p}_{i,l} = \textbf{W}_{i,j,k,l}\textbf{q}_{i,j,k},
    \label{eq-w}
\end{equation}
where

\begin{equation}
    \textbf{W}_{i,j,k,l}^T=\frac{1}{4A_{i,j,k}^2}\left(\begin{array}{c}
    -D(i,j,k;i,k,l)\textbf{I}\\
    D(i,j,k;i,j,l)\textbf{I}\\
    6V_{i,j,k,l}\textbf{I}
    \end{array}\right),
    \label{eq-w2}
\end{equation}
with \textbf{I} being the $3\times3$ identity matrix.

\section{Workspace Definition}
\begin{figure}[t!]
    \centering
    \includegraphics[width=\columnwidth]{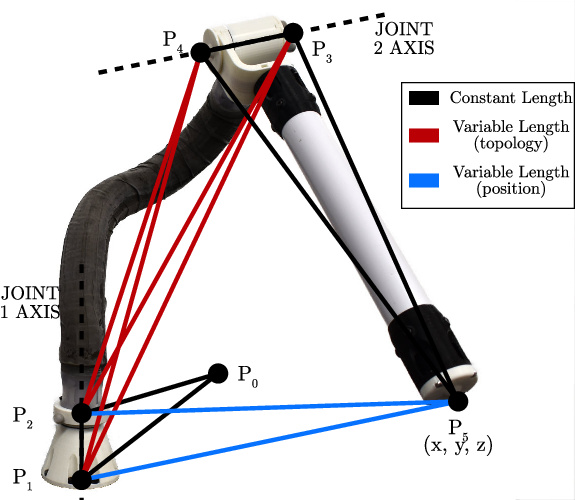}
    \caption{The 2-DOF malleable robot arm can be modelled as a bar-and-joint framework formed by connecting 5 points: $P_1$ and $P_2$, which define the first axis; $P_3$ and $P_4$, which define the second axis; and $P_5$, which corresponds to the centre of the end effector. An additional offset point $P_0$ is added, which is used to define the angle of the first axis. The distances between points can then be categorised as constant distances that do not change (\textbf{black}), distances that vary based on the topology configuration of the robot (\textbf{red}), and distances that vary based on the joint positioning of the robot (\textbf{blue}).}
    \label{distancegeometry}
\end{figure}

Our 2-DOF malleable robot is defined by two joints and a malleable link: a vertical revolute joint at the base, a malleable link connected co-linearly to this joint output, and a second revolute joint connected perpendicularly to the other end of the malleable link. Attached to this second joint is a rigid link, also attached perpendicularly, which then terminates at an end effector. We can model an arbitrary link connecting two skew revolute axes (in this case the malleable link) as a tetrahedron by selecting two points along each of the joint axes and connecting them all with edges, and we can model the rigid link connected to the secondary revolute axis as a triangle by connecting the two joint axis points and a single point at the end effector similarly \cite{rojas-coupler}. We can then model the developed 2-DOF malleable robot using distance geometry as a bar-and-joint framework of 6 points and 12 edges, shown in Fig.~\ref{distancegeometry}, with $P_5$ corresponding to the end-effector, and $P_1$ corresponding to the robot origin. The axes of the revolute joints are defined by the points $P_1$ and $P_2$ for joint 1, and $P_3$ and $P_4$ for joint 2. An additional fixed point $P_0$ is defined offset from the origin at $P_1$, which is necessary for the forward and inverse kinematics, but not for the workspace definition or topology reconfiguration. The interpoint distances can further be categorised as:
\begin{enumerate}
  \item Distances with constant length, that do not change with variation in robot topology and positioning ($\mathbf{p}_{0,1}$, $\mathbf{p}_{0,2}$, $\mathbf{p}_{1,2}$, $\mathbf{p}_{3,4}$, $\mathbf{p}_{3,5}$, $\mathbf{p}_{4,5}$).
  \item Distances that vary with changes in robot topology, but not robot positioning ($\mathbf{p}_{1,3}$, $\mathbf{p}_{1,4}$, $\mathbf{p}_{2,3}$, $\mathbf{p}_{2,4}$).
  \item Distances that vary with changes in robot positioning, but not robot topology ($\mathbf{p}_{1,5}$, $\mathbf{p}_{2,5}$).
\end{enumerate}

For defining the workspace, we can represent this as a Cayley-Menger determinant of 5 points (not including $P_0$), shown in Eq.~\ref{eq-CMD}. As $D(1,2,3,4,5) = D(4,3,2,1,5) = 0$, we can use the properties of the determinant of block matrices \cite{Powell} to show that this condition can be compactly expressed using $3\times3$ matrices as
\begin{align}
    D(1,2,3,4,5) = 2\,s_{1,2}\,s_{1,5}\,s_{2,5}\det(\mathbf{A}-\mathbf{B}\mathbf{C}\mathbf{B}^T)=0, \label{eq:determinant2}
\end{align}
where
\begin{align*}
    \mathbf{A} &= \left[ \begin{array}{ccc}
                0   &   1         & 1       \\
                1   &   0         & s_{3,4} \\
                1   &   s_{3,4}   & 0       \\
                \end{array} \right],\, 
    \mathbf{B} = \left[ \begin{array}{ccc}
                1   &   1         & 1       \\
                s_{2,4}   &   s_{1,4}         & s_{4,5} \\
                 s_{2,3}   &   s_{1,3}         & s_{3,5} \\
                \end{array} \right], \textrm{ and} \\           
    \mathbf{C} &= \frac{1}{2}\left[ \begin{array}{ccc}
                -\frac{s_{1,5}}{s_{1,2}\,s_{2,5}}   &  \frac{1}{s_{1,2}}         &    \frac{1}{s_{2,5}}    \\
                \frac{1}{s_{1,2}}   &   -\frac{s_{2,5}}{s_{1,2}\,s_{1,5}}         & \frac{1}{s_{1,5}} \\
                 \frac{1}{s_{2,5}}   &   \frac{1}{s_{1,5}}         & -\frac{s_{1,2}}{s_{1,5}\,s_{2,5}} \\
                \end{array} \right].
\end{align*}

\begin{figure*}[t!]
    \centering
    \includegraphics[width=\textwidth]{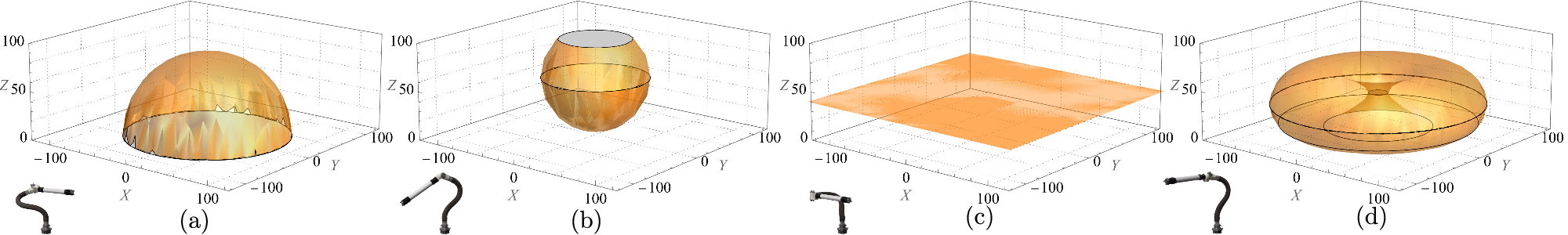}
    \caption{Generated theoretical example workspaces of a 2-DOF malleable robot for each of the robot topology types, using the equations defined in section IV which do not take into account joint limits. \textbf{(a)} Spherical, \textbf{(b)} PUMA-like, \textbf{(c)} SCARA, and \textbf{(d)} General Articulated.}
    \label{simulationworkspace}
\end{figure*}

Using the notation described in Fig.~\ref{distancegeometry}, we observe that in $\mathbb{E}^{3}$ (where a 2-DOF malleable robot can physically exist), equation \eqref{eq:determinant2} is solely satisfied. We can exploit this fact to compute the Cartesian equation of the robot workspace, say $\Gamma(x, y, z)$, by deriving the locus of point $P_5$ (the end effector), whose coordinates are $\mathbf{p}_5 = (x, y, z)$ in a particular reference frame. This computation can then be simplified by assuming, without loss of generality, that $P_1$ is equal to the origin of the global reference frame and that $P_2$ is located in the positive side of the $z$-axis, such that $\mathbf{p}_1 = (0, 0, 0)$ and $\mathbf{p}_2 = (0, 0, d_{1,2})$. Consequently,
\begin{align}
\begin{split}
    s_{1,5} &= x^2 + y^2 + z^2\\
    s_{2,5} &= x^2 + y^2 + z^2 - 2d_{1,2}z + s_{1,2}. \label{eq:basedistances}
\end{split}
\end{align}

Substituting equation \eqref{eq:basedistances} into equation \eqref{eq:determinant2} and fully expanding the result and rearranging terms, we obtain
\begin{align}
    \Gamma(x, y, z) & \myeq  q_0\,(x^2+y^2+z^2)^2+q_1\,d_{1,2}\,z\,(x^2+y^2+z^2) \nonumber \\
    & q_2\,x^2+q_2\,y^2+q_3\,z^2+q_4\,d_{1,2}\,z+q_5, \label{eq:Gamma}
\end{align}
where $q_i,\,i=0,\ldots,5$ are polynomials in $s_{1,2}=d_{1,2}^2$, $s_{1,3}$, $s_{1,4}$, $s_{2,3}$, $s_{2,4}$, $s_{3,4}$, $s_{3,5}$, and $s_{4,5}$. $\Gamma(x, y, z)$ is a degree 4 (quartic) algebraic surface that corresponds to the workspace surface traced by the end effector (point $P_5$) of a 2-DOF malleable robot. The full expressions of the polynomials $q_i$ cannot be included here due to space limitations; however these polynomials can be easily reproduced using a computer algebra system following the above steps.

Additionally, we can define certain workspace categories belonging to specific robot configurations (topologies), by providing further constraints to the two revolute axes of the malleable robot. Malleable robots are designed to be general purpose serial robot manipulators, and so follow a similar approach where the task workspace defines the configuration. The robot configurations defined herein are Spherical, PUMA-like, SCARA, and General Articulated, with the constraint definitions of each discussed next.

\subsection{Spherical (or variable radius) case}
For the spherical robot configuration, the two revolute axes of the robot coincide at the base (in this case, the origin), such that, according to the notation of Fig.~\ref{distancegeometry}, points $P_1$ and $P_3$ are coincident. Thus, $s_{1,3} = 0$, $s_{2,3} = s_{1,2}$, and $s_{3,4} = s_{1,4}$. Substituting into \eqref{eq:Gamma} we get
\begin{align}
    \Gamma_A(x, y, z) \myeq x^2 + y^2 + z^2 - s_{3,5}=0,
\end{align}
which corresponds to the equation of a sphere of radius $d_{3,5}$ centred around $P_1$. Note that in this case the radius $d_{3,5}$ is not constant, and can be adjusted according to desired task. An example of this workspace can be seen in Fig.~\ref{simulationworkspace}(a).

\subsection{PUMA-like (or variable centre and radius) case}
For the PUMA-like robot configuration, the two revolute axes of the robot are perpendicular and coincide at a point located above the base (in the positive side of the $z$-axis), such that points $P_2$ and $P_4$ are coincident, and the angle $\angle P_1P_2P_3$ is $\frac{\pi}{2}$. Thus, $s_{2,4} = 0$, $s_{1,4} = s_{1,2}$, $s_{3,4} = s_{2,3}$, and $s_{1,3} = s_{1,2}+s_{2,3}$. Substituting these values into \eqref{eq:Gamma}, we obtain
\begin{align}
    \Gamma_B(x, y, z) \myeq x^2 + y^2 + (z-d_{1,2})^2-s_{4,5}=0,
\end{align}
which corresponds to the equation of a sphere of radius $d_{4,5}$ centred at $P_2$. In this case, both the centre $(0, 0, d_{1,2})$ and radius $d_{4,5}$ are not constant, they can be adjusted according to the required task. Note that the same equation is obtained when the perpendicularity of the two axes is relaxed. An example of such a workspace can be seen in Fig.~\ref{simulationworkspace}(b).

\subsection{SCARA (or planar) case}
For a SCARA robot configuration, the two revolute axes of the robot are parallel, resulting in a planar workspace. Using projective geometry arguments, this implies that there exist a point in the second axis, say $P_3$, such that the distance between it and the $xy$-plane is $\delta$, with $\delta>0$, $\delta \to \infty$. Hence, $d_{1,3}=\delta$, $d_{2,3}=d_{1,2}+\delta$, $d_{3,4}=z_{4}+\delta$, $d_{3,5}=z_{5}+\delta$, where $z_{i}$ is the distance between $P_i$ and the $xy$-plane. Substituting these values into \eqref{eq:Gamma}, we obtain an equation that can be arranged as a quadratic polynomial in $\delta$: $\Omega = k_2(x,y,z)\delta^2+k_1(x,y,z)\delta+k_0(x,y,z) = 0$. By factoring $\delta^2$ out of this polynomial, we obtain $\Omega =\delta^2(k_2(x,y,z)+\frac{k_1(x,y,z)}{\delta}+\frac{k_0(x,y,z)}{\delta})=0$. Since $\delta \to \infty$, then $\Omega = k_2(x,y,z)=0$. 

As the two revolute axes of the robot are parallel, it is necessary to include additional constraints in $\Omega=k_2(x,y,z)=0$, that is, $P_2$=$P_4$=$P^\infty$. This implies that $s_{2, 4} = 0$ and $d_{1, 4} = d_{1, 2}$. Substituting these values into $\Omega=k_2(x,y,z)=0$, we get $(z_4-d_{1, 2})\,s_{1, 2}\,\Phi(x,y,z) = \Phi(x,y,z) = 0$. Finally, we can include the constraint $z_4 = d_{1, 2}$ (as $P_2$=$P_4$) in the result ($\Phi(x,y,z)$), yielding,
\begin{equation*}
     \left( z-{\it z_5} \right)  \left( {x}^{2}+{y}^{2}+{z}^{2}-2\,d_{{1,2}
}z+{d_{{1,2}}}^{2}-s_{{4,5}} \right) = 0.
\end{equation*}
We can then follow a similar procedure to that for $\delta$, but instead for $d_{1,2}$ ($d_{1,2} \to \infty$ since $P_2=P_4=P^\infty$), and finally we obtain
\begin{align}
    \Gamma_C(x, y, z) \myeq \left( z-{\it z_5} \right)=0, \label{eq:GammaSCARA}
\end{align}
which corresponds to the equation of a plane parallel to the $xy$-plane. Observe that $z_5$, the distance between the end effector and the $xy$-plane, is not constant and can be adjusted according to need. An example of this workspace can be seen in Fig.~\ref{simulationworkspace}(c).

\subsection{General articulated}
Finally, we define the general articulated robot configuration as any robot configuration that does not satisfy any of the constraints of the 3 previously defined robot configurations. Thus, the structure of this workspace surface is $\Gamma(x, y, z)=0$ (equation \eqref{eq:Gamma}). An example of this workspace, corresponding to a torus, can be seen in Fig.~\ref{simulationworkspace}(d).

\section{Forward and Inverse Kinematics}
\subsection{Forward Kinematics}
In computing the forward kinematics of the malleable robot, we specify the joint angles (in this case, the dihedral angles--use of the physical joint values is addressed later), and obtain the new position of $P_5$.

We assume the robot is in a fixed topology, and that the current point positions (and therefore their interpoint distances) are known. Starting from the origin, we define $\phi_{1,0,2,3}$ as the joint 1 dihedral angle between the fixed triangle $\ppp_{0,1,2}$ and the current topology defined triangle $\ppp_{1,2,3}$. We then define $\phi_{3,2,4,5}$ as the joint 2 dihedral angle between the current topology defined triangle $\ppp_{2,3,4}$ and the constant length triangle $\ppp_{3,4,5}$.

We first compute the new location of P3, defined by the new dihedral angle $\phi_{1,0,2,3}$ value. Using the interpoint distances we can calculate the areas of both triangles ($A_{0,1,2}$,$A_{1,2,3}$) using Eq.~\ref{eq1}. Substituting into Eq.~\ref{eq3}, we can solve for the new distance $\mathbf{p}_{0,3}$. With all distances for the tetrahedron $\pppp_{1,0,2,3}$ known, we can compute the new position of $P_3$ using trilateration.

After using Eq.~\ref{eq2} to compute the orientated volume $V_{1,0,2,3}$, we can use the now known tetrahedron $\pppp_{1,0,2,3}$ distances to compute the new position of point $P_3$:

\begin{equation}
    P_3 = P_1+\textbf{W}_{1,0,2,3}\textbf{q}_{1,0,2}.
    \label{eq-p3}
\end{equation}

With the new position of $P_3$ known, and the constant distance $\textbf{p}_{3,4}$, we can compute the position of $P_4$ using the points $P_1$, $P_2$, and $P_3$ and their known interpoint distances:

\begin{equation}
    P_4 = P_1+\textbf{W}_{1,3,2,4}\textbf{q}_{1,3,2}.
    \label{eq-p4}
\end{equation}

Finally, we can compute the new location of $P_5$ using the same process for $P_3$, using the new positions of $P_3$ and $P_4$ and the new dihedral angle $\phi_{3,2,4,5}$ for joint 2:

\begin{equation}
    P_5 = P_3+\textbf{W}_{3,2,4,5}\textbf{q}_{3,2,4}.
    \label{eq-p5}
\end{equation}

As the orientation of the volume of tetrahedra requires all distances to be known (Eq.~\ref{eq2}), in determining $P_3$ and $P_5$ we instead define the orientation based on the dihedral angles, where it is defined as positive if $\phi_{i,j,k,l}<180^{\circ}$ and negative otherwise, where $0^{\circ}\leq\phi_{i,j,k,l}\leq360^{\circ}$.

\subsection{Inverse Kinematics}
For the inverse kinematics of the robot, we provide the end effector position ($P_5$) and compute the required dihedral angles necessary to obtain it. It is assumed the current topology of the robot (its positions and interpoint distances) are all known.

We first calculate the angle of joint 2, $\phi_{3,2,4,5}$. Using the new position of $P_5$, we know the new distance $\textbf{p}_{2,5}$. Rearranging Eq.~\ref{eq3} to solve for $\phi_{i,j,k,l}$ we obtain

\begin{align}
    \cos{(\phi_{i,j,k,l})}&=\frac{D(i,j,k;i,k,l)}{4A_{i,j,k}A_{i,k,l}},
    \nonumber \\
    &=\frac{D(i,j,k;i,k,l)}{D^\frac{1}{2}(i,j,k)D^\frac{1}{2}(i,k,l)}.
    \label{eq-phi2}
\end{align}

\begin{figure}[t!]
    \centering
    \includegraphics[width=\columnwidth]{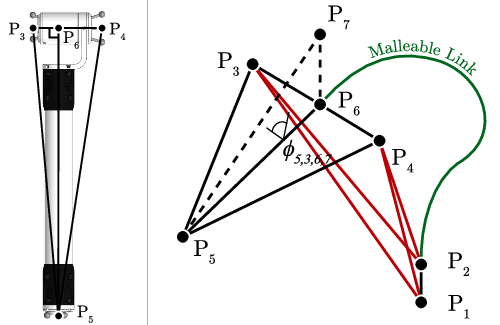}
    \caption{Bar-and-joint framework of the malleable robot with the desired end-effector location, $P_{5}$, with vector $p_{5,6}$ defined by the point $P_{6}$ that defines the end effector orientation. Known distances within the robot are shown in \textbf{black}, and unknown distances (that must be computed with each configuration) are shown in \textbf{red}. A \textbf{green} distance depicts the malleable link. By varying the value of the dihedral angle $\phi_{5,3,6,7}$, alternative positions of points $P_3$ and $P_4$ can be generated for a specific end effector position and orientation. The relation of the points to the physical robot is shown on the left.}
    \label{distances}
\end{figure}

Using Eq.~\ref{eq-phi2} along with Eq.~\ref{eq-tri-area}, we can calculate the angle of joint 2 $\phi_{3,2,4,5}$ using only interpoint distances. This procedure returns the smallest value of the dihedral angle, limited to the range $0^{\circ}\leq\phi_{i,j,k,l}\leq180^{\circ}$. We can extend this value to the full $360^{\circ}$ range to find the two valid values of the dihedral angle, where it is $360^{\circ}-\phi_{i,j,k,l}$ or $\phi_{i,j,k,l}$ otherwise. These are known as the \textit{elbow up} and \textit{elbow down} configurations of an arm.

Next, we can compute the new position of $P_3$ as performed in the forward kinematics using Eq.~\ref{eq-w}, using the current topology points of the robot, along with the new position of $P_5$. Thus,

\begin{equation}
    P_3 = P_5+\textbf{W}_{5,1,2,3}\textbf{q}_{5,1,2}.
    \label{eq-ik-p3}
\end{equation}
With the new position of $P_3$ known, we can repeat the procedure for calculating $\phi_{3,2,4,5}$ for $\phi_{1,0,2,3}$. If desired, the new location of $P_4$ can also be computed using Eq.~\ref{eq-w} and the calculated dihedral angle $\phi_{1,0,2,3}$.

\begin{figure}[t!]
    \centering
    \includegraphics[width=0.8\columnwidth]{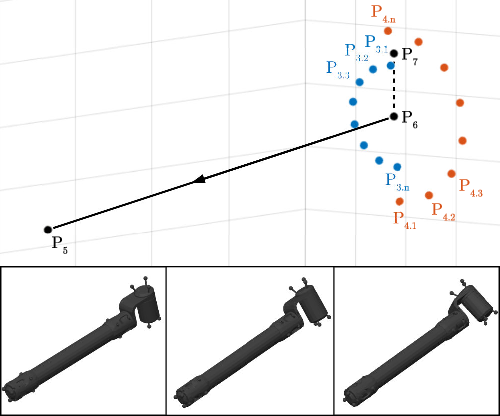}
    \caption{\textbf{Top:} A generated sampling ($n=8$) of points $P_3$ and $P_4$ (simulated) from a given end effector specification (experimental measurement). \textbf{Bottom:} Simulated distal link orientations for $n=2$, $n=3$, and $n=4$ are shown.}
    \label{pointsSampler}
\end{figure}

\section{Topology Reconfiguration}
The key advantage of malleable robots over traditional robots is the ability to reconfigure their topology, resulting in a change in their relative revolute joint positions and therefore a change in the working environment of the robot. Despite the low DOF of the 2-DOF malleable robot, this reconfiguration enables a much larger working environment. To fully utilise this advantage, we must determine the necessary reconfiguration required for a desired end effector position, allowing us to dynamically reconfigure the topology of the malleable robot, the relation between the joint axes and the distances between them, depending on the task workspace required.

\begin{figure*}[t!]
    \centering
    \includegraphics[width=\textwidth]{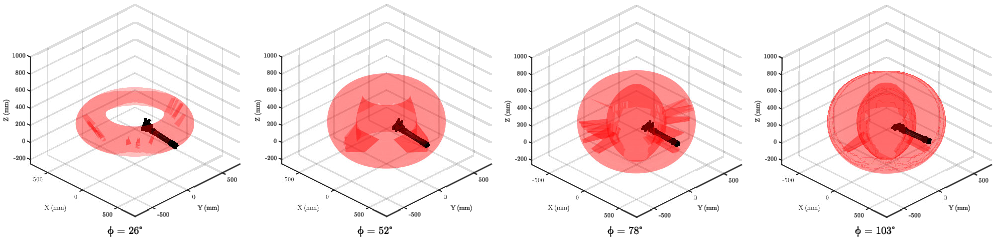}
    \caption{Simulated workspaces of a single topology reconfiguration. The distal linkage positions of each configuration along with their resulting workspaces are shown. Below each subfigure the value of the dihedral angle $\phi$ of the tetrahedron $\pppp_{5,3,6,7}$ is listed.}
    \label{workspace}
\end{figure*}

When computing the forward and inverse kinematics, we assume the geometric topology of robot is known. To compute the topology reconfiguration, we instead assume only the constant distances and the distances that define the end effector position (variable length--position) are known, shown in Fig.~\ref{distancegeometry}. To compute a robot topology, a single desired end effector position along with the partial orientation of the distal link is required. We assume this point is reachable, and thus exists within the reconfiguration workspace of the robot. We define the end effector position as $P_5$, with the orientation of the distal link determined by a second point $P_6$ located along the central axis of the distal link, perpendicular to the vector $\textbf{p}_{3,4}$, as shown in Fig.~\ref{distances}~(left). The orientation of the distal link is therefore along the vector $\textbf{p}_{6,5}$, however is not restricted in rotation about this vector.

\begin{figure}[t!]
    \centering
    \includegraphics[width=\columnwidth]{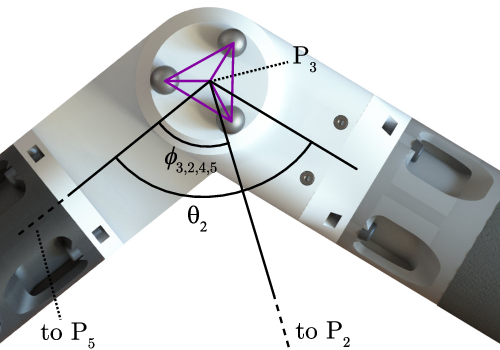}
    \caption{Demonstration of the joint offset problem, overlaid a joint 2 of the malleable robot. The joint absolute value (\textbf{$\theta_2$}) is shown to differ from the plane angle of joint 2 (\textbf{$\phi_{3,2,4,5}$}).}
    \label{jointoffset}
\end{figure}

\begin{figure}[t!]
    \centering
    \includegraphics[width=\columnwidth]{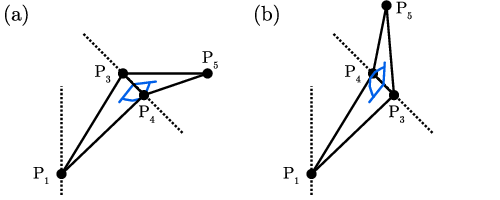}
    \caption{Demonstration of two reconfigurations which produce identical workspaces and have identical dihedral angles, however have different joint offset directions. \textbf{(a)} A reconfiguration with a positive joint offset and \textbf{(b)} a reconfiguration with a negative joint offset.}
    \label{jointPredict}
\end{figure}

We sample (interpolate discrete points) a desired number of orientations, $n$, of the distal link around the axis $\textbf{p}_{6,5}$, which produces a set of reconfigurations and pairing workspaces, of which an optimal workspace can then be selected by the user based on their desired robot topology. Given a desired $P_5$ and $P_6$, we first define a point $P_7$ in the positive z-axis above $P_6$. With this new $P_7$, we sample the dihedral angle $\phi_{5,3i,6,7}=\frac{\pi(i-1)}{n-1}$ where $1\leq i\leq n$ is the current sample. The bar-joint framework of the components of this method is shown in Fig.~\ref{distances}~(right). From the sampled dihedral angle $\phi_{5,3i,6,7}$, we first compute the new distance $s_{3,7_i}$ using eq.~\ref{eq-phi2}, solving for the unknown distance. With all distances then known in the tetrahedron $\pppp_{5,3i,6,7}$, we compute the position of $P_{3i}$ for each sampling as

\begin{equation}
    P_{3i} = P_5+\textbf{W}_{5,7,6,3i}\textbf{q}_{5,7,6}.
\end{equation}

With the position of $P_{3i}$ known for each sample, we can compute the position of $P_{4i}$ using the tetrahedron $\pppp_{5,3i,6,4i}$

\begin{equation}
    P_{4i} = P_5+\textbf{W}_{5,3i,6,4i}\textbf{q}_{5,3i,6},
\end{equation}
where the distance $d_{3i,4i}=d_{3i,6}+d_{4i,6}$, resulting in a flattened tetrahedron with a dihedral angle of $\pi$. An example sampling demonstrating the generation of points $P_{3i}$ and $P_{4i}$ for a desired end effector position is shown in Fig.~\ref{pointsSampler}. Simulated workspaces along with their distal link orientation for a subset of these examples are shown in Fig.~\ref{workspace}.

\begin{figure}[t!]
    \centering
    \includegraphics[width=\columnwidth]{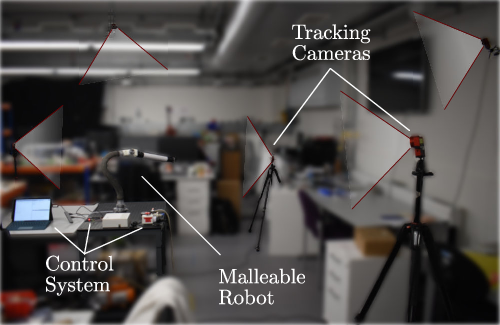}
    \caption{Experimental setup for motion tracking the malleable robot.}
    \label{experimentsetup}
\end{figure}

With the locations of $P_1$ to $P_5$ for each sample known, we can simply obtain the interpoint distances (specifically $d_{1,3}$, $d_{2,3}$, $d_{1,4}$, $d_{2,4}$) which define the robot topology.

\begin{figure*}[t!]
    \centering
    \includegraphics[width=0.99\textwidth]{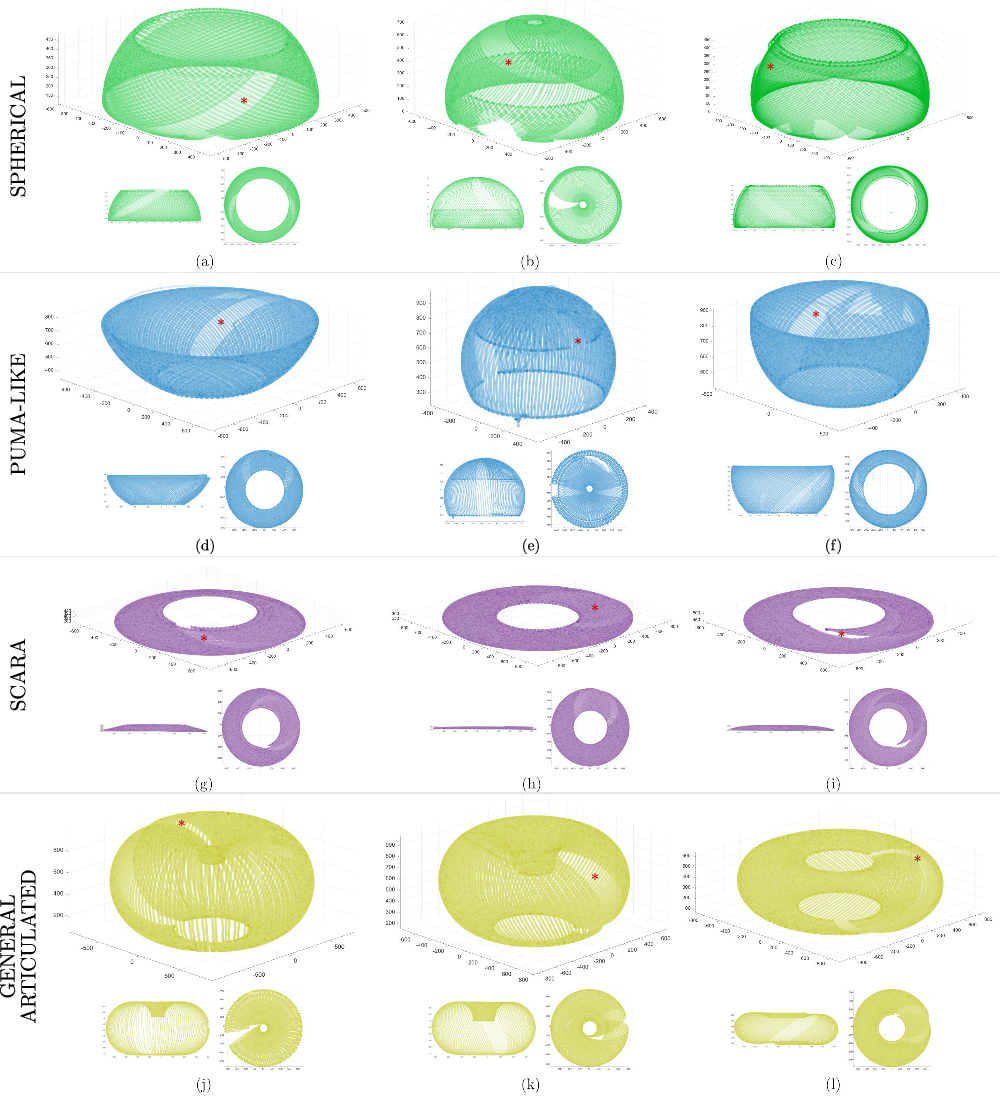}
    \caption{Experimental results of the workspaces of the 2-DOF malleable robot, for 3 configurations of each type of robot topology, (Top: ISO; Bottom left to right: YZ and XY): (\textbf{a})(\textbf{b})(\textbf{c}) Spherical (green), (\textbf{d})(\textbf{e})(\textbf{f}) PUMA-like (blue), (\textbf{g})(\textbf{h})(\textbf{i}) SCARA (purple), and (\textbf{j})(\textbf{k})(\textbf{l}) General Articulated (yellow). Areas of the workspaces missing due to joint limits are highlighted with a red asterisk. Constant distances (mm) across all configurations were $d_{1,2}$ = 58, $d_{3,4}$ = 49, $d_{3,5}$ = 455, and $d_{4,5}$ = 460. Specific distances (mm) for each configuration were (\textbf{a}) $d_{1,3}$ = 395, $d_{1,4}$ = 444, $d_{1,5}$ = 614, $d_{2,3}$ = 343, $d_{2,4}$ = 392, $d_{2,5}$ = 572, (\textbf{b}) $d_{1,3}$ = 511, $d_{1,4}$ = 559, $d_{1,5}$ = 695, $d_{2,3}$ = 472, $d_{2,4}$ = 520, $d_{2,5}$ = 666, (\textbf{c}) $d_{1,3}$ = 349, $d_{1,4}$ = 396, $d_{1,5}$ = 443, $d_{2,3}$ = 299, $d_{2,4}$ = 347, $d_{2,5}$ = 441, (\textbf{d}) $d_{1,3}$ = 650, $d_{1,4}$ = 667, $d_{1,5}$ = 969, $d_{2,3}$ = 603, $d_{2,4}$ = 619, $d_{2,5}$ = 930, (\textbf{e}) $d_{1,3}$ = 523, $d_{1,4}$ = 525, $d_{1,5}$ = 529, $d_{2,3}$ = 466, $d_{2,4}$ = 469, $d_{2,5}$ = 497, (\textbf{f}) $d_{1,3}$ = 642, $d_{1,4}$ = 669, $d_{1,5}$ = 551, $d_{2,3}$ = 588, $d_{2,4}$ = 614, $d_{2,5}$ = 510, (\textbf{g}) $d_{1,3}$ = 500, $d_{1,4}$ = 544, $d_{1,5}$ = 562, $d_{2,3}$ = 451, $d_{2,4}$ = 494, $d_{2,5}$ = 520, (\textbf{h}) $d_{1,3}$ = 569, $d_{1,4}$ = 596, $d_{1,5}$ = 503, $d_{2,3}$ = 540, $d_{2,4}$ = 564, $d_{2,5}$ = 474, (\textbf{i}) $d_{1,3}$ = 535, $d_{1,4}$ = 581, $d_{1,5}$ = 647, $d_{2,3}$ = 483, $d_{2,4}$ = 528, $d_{2,5}$ = 604, (\textbf{j}) $d_{1,3}$ = 671, $d_{1,4}$ = 670, $d_{1,5}$ = 1009, $d_{2,3}$ = 626, $d_{2,4}$ = 625, $d_{2,5}$ = 981, (\textbf{k}) $d_{1,3}$ = 668, $d_{1,4}$ = 656, $d_{1,5}$ = 385, $d_{2,3}$ = 620, $d_{2,4}$ = 609, $d_{2,5}$ = 3647, (\textbf{l}) $d_{1,3}$ = 593, $d_{1,4}$ = 618, $d_{1,5}$ = 726, $d_{2,3}$ = 558, $d_{2,4}$ = 581, $d_{2,5}$ = 718.}
    \label{expresults}
\end{figure*}

\begin{figure*}[t!]
    \centering
    \includegraphics[width=\textwidth]{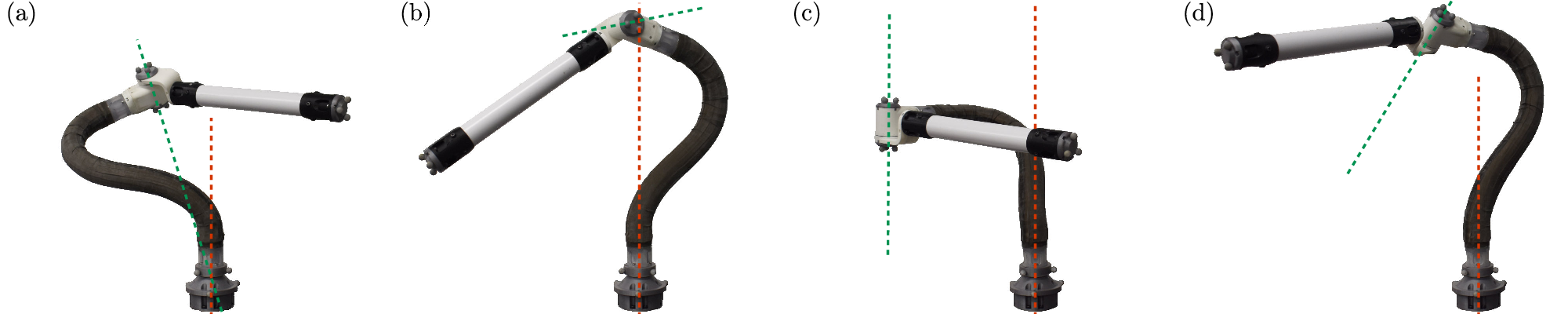}
    \caption{The defined workspace categories of the malleable robot topologies: \textbf{(a)} Spherical, \textbf{(b)} PUMA-like, \textbf{(c)} SCARA, and \textbf{(d)} General Articulated.}
    \label{malleablerobotconfiguration}
\end{figure*}

\begin{figure*}[t!]
    \centering
    \includegraphics[width=\textwidth]{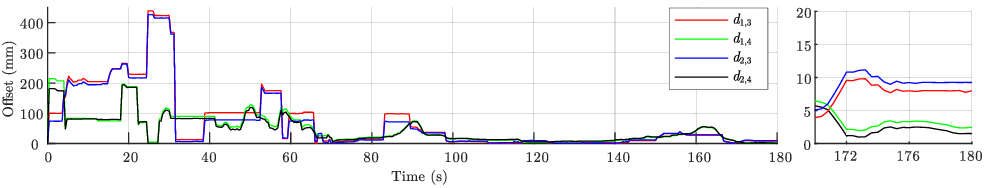}
    \caption{Alignment profile of the interpoint distances $d_{1,3}$, $d_{1,4}$, $d_{2,3}$, and $d_{2,4}$ of the malleable robot over time, as a manual reconfiguration is completed. An enlarged plot is shown far right of the final position, showing a $<$10~mm individual interpoint offset.}
    \label{alignment}
\end{figure*}

\subsection{Joint Recalibration}
Once a robot topology reconfiguration is carried out, a problem arises where the relation between the encoders of the joints and the dihedral angles of the joints change. This is due to a topology reconfiguration being determined only by $P_3$ and $P_4$, allowing free movement and positioning of the rest of the robot. As the kinematics of the robot is determined using the dihedral angles, and the control of the robot is achieved using the physical joint encoders, this relation needs to be determined after each topology reconfiguration. A demonstration of this joint offset problem can be seen in Fig.~\ref{jointoffset}, where a joint absolute value (\textbf{$\theta_2$}) differs from the dihedral angle of joint 2 (\textbf{$\phi_{3,2,4,5}$}).

It is assumed the topology of the reconfigured robot is known. From this topology we can compute the current dihedral angles, $\phi_{1,0,2,3}$ and $\phi_{3,2,4,5}$, using Eq.~\ref{eq3}. From the physical robot, we can also directly measure the current motor positions, $\theta_1$ and $\theta_2$. Finally, we must compute the direction of the motor alignment, which can be performed by moving each motor and measuring the change in angle. If it is positive, we define the offset direction coefficient, $k$, as 1, and -1 otherwise. We can then compute the offset amount, $\beta$, using the following

\begin{equation}
    \beta=\theta-k\phi_{i,j,k,l},
    \label{eq-ab}
\end{equation}
with known values of $k$ and $\beta$ for each joint, the robot is calibrated, and the kinematics of the robot can be directly converted to joint positions by rearranging Eq.~\ref{eq-ab} for $\theta$.

Considering the prediction of this joint offset, we observe that for a specific topology, there are two ‘valid’ reconfigurations that satisfy the geometry, which will produce identical workspaces. Both 'valid' reconfigurations are demonstrated in Figure \ref{jointPredict}. This is because, while the base joint is fixed, the second joint can align with the desired axis in either direction. The offset direction coefficient correlates with the dihedral angle $\phi_{3,2,4,5}$ orientation, which can be determined by computing the scalar triple product:

\begin{equation}
    P_3\cdot(P_4\times P_5),
    \label{stp}
\end{equation}

The variation in sign of the resulting scalar triple product directly correlates with the offset direction coefficient. When the scalar triple product is negative, so is the offset direction, and vice versa.

The offset amount $\beta$ is however more difficult to predict. The location where the malleable link enters and rigid link exits the secondary joint once reconfigured are not known, as it is up to the user how they would like to reconfigure the robot. In other words, the secondary joint is free to rotate around its axis before being fixed in place, as herein we are only interested in aligning the axes of the robot to the desired axes. As the angular position of the second link is then up to the user, it is not possible to predict the offset amount.

\section{Experimental Evaluation}

We first evaluated the end effector workspaces of the malleable robot in different robot topologies to demonstrate its capability to generate an infinite number of workspaces. Furthermore, this also allowed us to prove the viability of the malleable robot concept. Reflective motion tracking markers were fixed to each joint of the robot and the end effector. Seven OptiTrack Flex3 cameras were used to track the movement of the robot, and the experimental setup can be seen in Fig.~\ref{experimentsetup}. The calibration error of the cameras detailed a mean 3D error for overall projection as 0.455~mm and overall wand error as 0.081~mm. Prior to measurement, the desired geometric topology of the robot was selected. Using real-time feedback from the motion tracking, the malleable robot was manually configured and fixed in position. A total of twelve robot topology geometries were selected, 3 for each of the robot topology types defined in section IV.

The parameters (\emph{i.e.}, distances) that define the geometry of the robot for each of the configurations are given in Fig.~\ref{expresults}, alongside the collected experimental results. Once the geometry of the robot was confirmed and set, the tracking markers on each joint were removed leaving only a sole marker at the end effector. The robot then followed a predetermined path covering the entire workspace with the end effector by reaching all possible joint positions in steps of 0.088$^\circ$. Due to limits imposed on the joints, the maximum angles physically achievable by each joint were 10$^\circ$ to 350$^\circ$ for the primary joint and 53$^\circ$ to 307$^\circ$ for the secondary joint, with 0$^\circ$ aligned co-linearly with the connecting link. Examples for each of the robot topology types can be seen in Fig.~\ref{malleablerobotconfiguration}.

The geometric parameters of the robot (\emph{i.e.}, distances) for each assessed configuration are presented in Fig.~\ref{expresults}, where the experiment results obtained are shown. Once each geometry was confirmed, the tracking markers on the joints were removed, leaving only the single marker on the end effector. The end effector was then moved throughout the entire workspace of each robot topology by progressing through all possible joint positions in steps of 0.088$^\circ$. Due to joint limits, the maximum angles actually achievable by each joint were 10$^\circ$ to 350$^\circ$ and 53$^\circ$ to 307$^\circ$, for the primary and secondary joint, respectively, with 0$^\circ$ aligned with the connecting link. Examples of each robot topology type can be seen in Fig.~\ref{malleablerobotconfiguration}.

\begin{figure}[t!]
    \centering
    \includegraphics[width=\columnwidth]{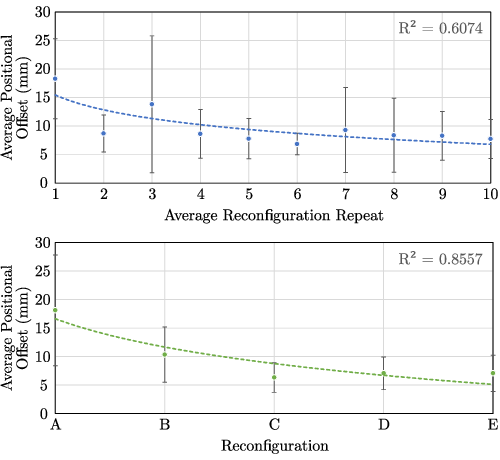}
    \caption{\textbf{Top:} Average positional offset across successive repeated topology reconfigurations. \textbf{Bottom:} Average positional offset for successive different topology reconfigurations.}
    \label{learning}
\end{figure}

We next evaluate the ability to align the robot to the desired optimal geometric topology. Using the same motion tracking setup, we constantly track the positions of $P1-4$ and compute the interpoint distances in real-time. By comparing these actual interpoint distances to the desired interpoint distances to determine their offset, we can determine how close a user is to correctly aligning the robot in the optimal topology. An example of this alignment plot over the course of a manual reconfiguration can be seen in Fig.~\ref{alignment}. A 3D plot of the desired and actual points is also available to the user in real time for aid in alignment. Once an alignment is completed, the robot is fixed into position and the final accuracy of the topology instance is measured. As it is not obvious to the user immediately how the robot should be reconfigured, we evaluated the attainable accuracy over 10 successive repeats of the desired reconfigurations, determining how knowledge and experience of a topology instance affected its attainable accuracy. A time limit of 3 minutes was given for each alignment to ensure consistency. This was repeated for 5 different geometric topologies. The results showing how the positional offset varies over a repeated topology reconfiguration, as well as how it varies over successive different topology reconfigurations, can be seen in Fig.~\ref{learning}.

\begin{figure*}[t!]
    \centering
    \includegraphics[width=\textwidth]{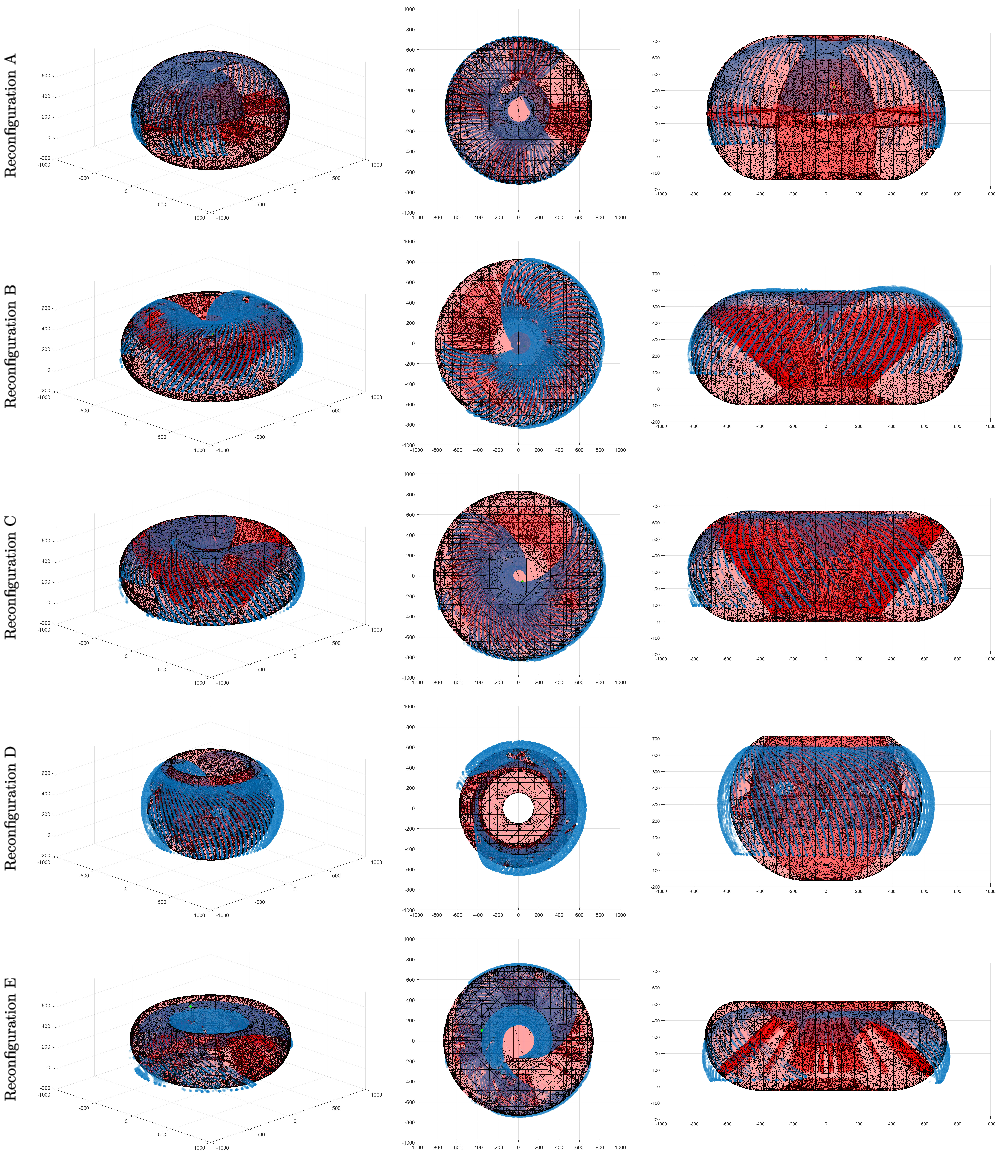}
    \caption{Experimental results showing measured workspaces of the 2-DOF malleable robot following each of the reconfigurations \textbf{(blue)}, overlaid with the theoretical modelled ideal workspace \textbf{(red)} and initial desired end effector location \textbf{(green)}. A random repeat is shown for each reconfiguration. Specific solutions for each reconfiguration (in mm) were: \textbf{A}: $P3 = [-165.38,233.04,282.09]$, $P4 = [-117.74,329.18,305.53]$, $P5 = [230,50,420]$, $P6 = [-150,270,290]$, $d_{1,3}=401.54$, $d_{2,3}=376.89$, $d_{1,4}=464.29$, $d_{2,4}=441.22$, \textbf{B}: $P3 = [-276.22,261.15,284.96]$, $P4 = [-201.81,308.06,219.37]$, $P5 = [60,-30,390]$, $P6 = [-250,280,260]$, $d_{1,3}=475.08$, $d_{2,3}=454.21$, $d_{1,4}=428.66$, $d_{2,4}=411.25$, \textbf{C}: $P3 = [-345.38,161.91,294.70]$, $P4 = [-307.37,239.70,362.00]$, $P5 = [40,-60,400]$, $P6 = [-330,190,320]$, $d_{1,3}=482.03$, $d_{2,3}=460.71$, $d_{1,4}=531.95$, $d_{2,4}=507.92$, \textbf{D}: $P3 = [29.61,318.51,264.36]$, $P4 = [61.37,216.33,288.35]$, $P5 = [-380,130,190]$, $P6 = [40,280,273]$, $d_{1,3}=414.98$, $d_{2,3}=392.83$, $d_{1,4}=365.66$, $d_{2,4}=337.74$, and \textbf{E}: $P3 = [-66.30,294.97,199.19]$, $P4 = [6.63,282.95,280.16]$, $P5 = [-360,100,490]$, $P6 = [-40,290,230]$, $d_{1,3}=362.05$, $d_{2,3}=343.40$, $d_{1,4}=398.34$, $d_{2,4}=373.56$.}
    \label{expworkspace}
\end{figure*}

We also evaluate the final reconfiguration accuracy of multiple topologies. Five geometric topologies were selected, and following manual reconfiguration into the new topologies each was evaluated in terms of interpoint distance offsets. This was repeated 5 times for each topology to obtain the average interpoint distances offset, and the results are shown in Table~\ref{results-accuracy-reconfiguration}, along with the overall average interpoint distance offset for each topology. These reconfigurations were each further evaluated in terms of their offset from their initial desired end effector points. Once reconfigured, the robot was instructed to explore its new workspace while the end-effector location was tracked. The resulting point-cloud workspace was then compared with the desired points to determine the distance between the nearest point-cloud point to the desired point, and its offset, as a method for determining if the desired point was within the reconfigured robot workspace. The results of this can be seen in Table~\ref{results-accuracy-alignment}, again showing the average of 5 repeats for each topology. The normal distance and average offset is also shown. A comparison of the achieved and desired workspaces for the 5 topologies can be seen in Fig.~\ref{expworkspace}.

Finally, we evaluate the payload of the developed malleable robot. The robot was reconfigured into example topologies of each of the 4 workspace categories, with P3 and P4 located roughly at their average distance from P1, and joint 2 at an angle of 45$^\circ$ from fully extended (actual value of 135$^\circ$). Slotted masses held via a hook were progressively added in increments of 50~g to the end effector at P5, and the positional offset of the end effector was measured using the motion tracking cameras. Results can be seen in Fig.~\ref{payload}.

\section{Discussion}
The results of the workspace exploration can be seen in Fig.~\ref{expresults}, with 3 different variations of each geometry within the same topology. Four distinct workspace categories were identified, correlating with the malleable robot topologies spherical, PUMA-like, SCARA, and general articulated. For the non-planar configurations (spherical, PUMA-like, and general articulated), the shape of the workspace demonstrated significant physical variation with the change of geometric parameters, producing a change in the radius of the resulting sphere (spherical [Fig.~\ref{expresults}(a)-(c)] and PUMA-like [Fig.~\ref{expresults}(d)-(f)] case), and flattened tori (general articulated case) [Fig.~\ref{expresults}(j)-(l)]. In the case of the planar SCARA configuration, variations in workspace height (z-axis) and internal radius only were observed, due to the fixed rigid link length defining the width of planar surface. A missing slice can be observed (highlighted in Fig.~\ref{expresults} with red asterisks) on each of the resulting workspaces, corresponding to areas not accessible by the motion tracking cameras and joint angle limits.

\begin{table}[!t]
    \caption{Reconfiguration accuracy results, showing the average variation in interpoint distances, $d_{1,3}$, $d_{1,4}$, $d_{2,3}$, and $d_{2,4}$, between the desired and achieved positioning of the distal link of the malleable robot. All dimensions are shown in mm.}
    \label{results-accuracy-reconfiguration}
    \centering
    \begin{tabularx}{\columnwidth}{C{2cm}|C{0.75cm}C{0.75cm}C{0.75cm}C{0.75cm}|C{1.2cm}}
        \hline
        Reconfiguration& $d_{1,3}$ & $d_{1,4}$ & $d_{2,3}$ & $d_{2,4}$ & Average\\
        \hline
        A, $\Delta$ & 4.24 & 4.13 & 4.30 & 4.13 & 4.20\\
        B, $\Delta$ & 3.34 & 4.55 & 3.13 & 4.87 & 3.97\\
        C, $\Delta$ & 2.40 & 2.77 & 2.64 & 2.87 & 2.67\\
        D, $\Delta$ & 3.69 & 4.85 & 3.94 & 4.78 & 4.31\\
        E, $\Delta$ & 5.62 & 8.11 & 6.21 & 7.68 & 6.90\\
        \hline
        Average & 3.86 & 4.88 & 4.05 & 4.87 & - \\
        \hline
    \end{tabularx}
\end{table}

\begin{table}[!t]
    \caption{Alignment accuracy results, showing the average positional difference between the desired and actual robot end effector position. All dimensions are shown in mm.}
    \label{results-accuracy-alignment}
    \centering
    \begin{tabularx}{\columnwidth}{C{1.45cm}|C{0.65cm}C{0.65cm}C{0.65cm}|C{0.9cm}C{0.95cm}|C{0.6cm}}
        \hline
        Reconfiguration & $\Delta_X$ & $\Delta_Y$ & $\Delta_Z$ & Normal & Average & \% Error\\
        \hline
        A &  7.56 &  6.41 &  9.09 & 16.44 &  9.87 & 1.5\%\\
        B & 17.46 & 11.18 & 11.51 & 25.00 & 13.82 & 2.3\%\\
        C &  7.93 & 12.71 &  5.89 & 17.45 &  8.85 & 1.6\%\\
        D &  8.79 &  5.30 &  1.53 & 11.61 &  5.21 & 1.1\%\\
        E &  8.22 &  5.32 & 25.35 & 27.59 & 15.75 & 2.5\%\\
        \hline
        Average & 9.99 & 8.18 & 10.67 & 19.62 & - & 1.8\%\\
        \hline
    \end{tabularx}
\end{table}

By considering the construction and operation of the malleable robot we can justify some of the issues with the generated SCARA workspace. Ideally, the SCARA workspace should be a planar surface (as defined by equation \eqref{eq:GammaSCARA}), with the only variation of the end effector in the X and Y axes. However, in the experimental results a slight variation in height across the workspace can be observed, which can most likely be explained due to a misalignment between the joint axes of the malleable robot. Furthermore, when undergoing motion the centre of mass of the robot expectedly changes. Due to imperfect tolerances within the joints of the robot, specifically in the primary base joint, and wrinkles in the malleable link latex membrane, a slight variation in the desired end effector position can be seen as the robot progresses through the joint workspace. While this variation is minimal, it is an aspect that must be considered in the production of reconfigurable malleable robots. Following a completed motion of the robot, it was also noted that a minor oscillation of the malleable link (not-dissimilar from a spring) occurred before dissipating. This is again an aspect of malleable robots that must be considered, and was attributed to the elastic nature of the membranes in the malleable link, resulting in a slight elastic response as described in previous research on malleable links \cite{clark2019assessing}. Therefore, to reduce the effect of this, resulting in minimal variation and oscillation of the malleable link during an experiment, the joint speeds were reduced (25RPM) preventing extreme forces under directional changes.

\begin{figure}[t!]
    \centering
    \includegraphics[width=\columnwidth]{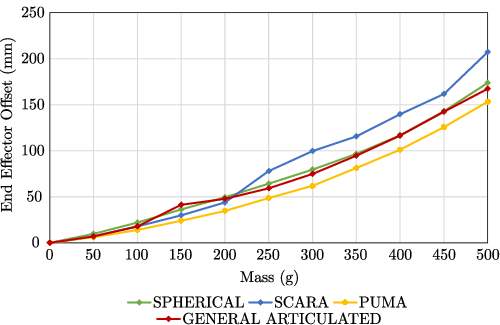}
    \caption{Plot of end-effector deflection against payload for the 4 different types of robot topology.}
    \label{payload}
\end{figure}

To conclude the workspace exploration results, we also identified that the manual (extrinsic) reconfiguration of the robot was challenging for a human to achieve accurately. For instance, in the PUMA-like case where a sphere was desired (Fig.~\ref{expresults}(f)), an overlapping sphere, a form of a minimal torus, was instead produced. This suggests that the second joint revolute axis did not perfectly intersect with the first as required, resulting in the alternative geometry of a general articulated case. Live-feedback to the user of the robot position from the motion tracking markers did allow for improved alignment of the joints when compared to pure visual alignment, however it did not account for the limitations of human manipulation accuracy in a 3D space. While the overall achieved accuracy was sufficient for our purposes, demonstrated by the majority of the measured workspaces in their correct desired form, it is clear that small variations in a reconfiguration can be critical to the accuracy of the resulting workspace.

We next consider aligning the malleable robot to a computed optimal geometric topology. From the results of successive repeated alignments (Fig.~\ref{learning}) we observe an improvement (logarithmic decay, $R^2=0.6074$) in the manual alignment with time, tending towards a consistently achievable average positional offset of 5-10~mm following around 5 repeats. When observed over successive different topology reconfigurations a similar trend was seen, with the addition of the average positional offset improving (logarithmic decay, $R^2=0.8557$) and the range of values for later configurations decreasing, shown by smaller error bars, suggesting user precision also improves with practice. From these results we can clearly state that aligning the malleable robot manually is not trivial, and requires practice and training to obtain a respectable alignment accuracy ($<$10~mm). Following this experiment, all following reconfigurations that were carried out were practised with a minimum of 10 repetitions to ensure the user was trained on the reconfiguration and that a low reconfiguration offset was achieved.

\begin{figure}[t!]
    \centering
    \includegraphics[width=\columnwidth]{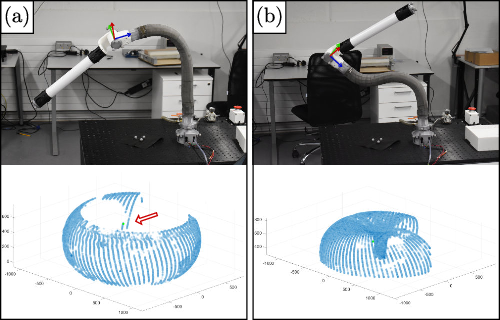}
    \caption{Reconfiguration A demonstrating how the orientation of the distal joint affects the reachable workspace. Configurations shown are both of the same geometric topology. \textbf{(a)} A configuration unable to reach the desired end effector location \textbf{(green)}, \textbf{(b)} a configuration capable of reaching the desired position.}
    \label{topologyVariation}
\end{figure}

\begin{figure}[t!]
    \centering
    \includegraphics[width=\columnwidth]{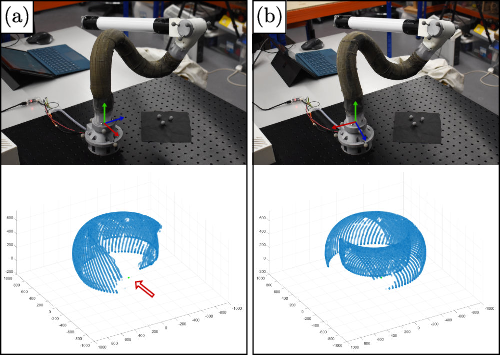}
    \caption{Reconfiguration D demonstrating how the orientation of the base joint affects the reachable workspace. Configurations shown are both of the same geometric topology. \textbf{(a)} A configuration unable to reach the desired end effector location \textbf{(green)}, \textbf{(b)} a configuration capable of reaching the desired position.}
    \label{jointVariation}
\end{figure}

The 5 robot topologies that were selected for assessing the reconfiguration accuracy, along with their measured and modelled workspaces, are detailed in Fig.~\ref{expworkspace}. The average alignment accuracy achieved, shown in Table~\ref{results-accuracy-reconfiguration}, for each of the reconfigurations was very high ($<$5~mm), with a maximum of 4.88~mm and minimum of 3.86~mm recorded for the variation in interpoint distances. The variation across the interpoint distances offsets was minimal, as expected due to the distances $d_{1,3}$ and $d_{2,3}$ being coupled and distances $d_{1,4}$ and $d_{2,4}$ coupled, with distances $d_{1,4}$ and $d_{2,4}$ showing a slightly higher average offset ($\sim$1~mm greater). Across the topologies the average offset was similar, with a maximum offset of 6.90~mm and minimum offset of 2.67~mm. It was observed that the reconfigurations that are closer to the origin (the base joint) of the malleable robot were easier in obtaining a small offset, due to the moment caused by the weight of the distal joint and link on the malleable link being reduced, and thus the deflection of the link was minimised. All reconfigurations were obtained within the time limit with ample time to spare, suggesting a successful reconfiguration could be achieved in 1-2~minutes, depending on the difficulty of the reconfiguration. It was determined an alignment of $<$10~mm offset was acceptable, however an ideal alignment was an offset of $<$5~mm. In aligning the robot, it was also determined to be key to achieve a consistent offset alignment across the interpoint distances (e.g. 5~mm, 5~mm, 5~mm, 5~mm), rather than obtaining a combination of high and low offsets (e.g. 10~mm, 0~mm, 10~mm, 0~mm), as similar offsets resulted in the desired workspace shape being achieved with minor translations, whereas a combination of offsets resulted in more drastic variations to the desired workspace. From the results obtained we can clearly state the robot was well aligned and ready for the following alignment accuracy experiment.

The results of the alignment accuracy experiment are shown in Table~\ref{results-accuracy-alignment}, and the experimental plot of one random repeat for each topology is shown overlaid the desired theoretical workspace in Fig.~\ref{expworkspace}. It was observed that the initial reconfiguration of the malleable robot required additional consideration due to the implemented joint angle limits. The joints were limited to prevent self collision and collision with the environment (such as the base platform), which were measured and applied following a robot reconfiguration and before each workspace exploration. Due to the joint limits, only a section of the theoretical workspace was achievable. Examples of this issue can be seen in Fig.~\ref{topologyVariation} and Fig.~\ref{jointVariation}, where the variation in distal joint position and base joint position are shown affecting the workspace, respectively, resulting in the desired end effector location being unobtainable. For the alignment accuracy experiment each reconfiguration was assessed and only reconfigurations that contained the desired end effector location were carried out.

The results of the alignment accuracy experiment show higher offsets than the reconfiguration accuracy results, with an average offset of the desired end effector location of 9.99~mm, 8.18~mm, and 10.67~mm for $\Delta_x$, $\Delta_y$, and $\Delta_z$, respectively. Combined, these axial offsets resulted in an average normal offset of 19.62~mm across all reconfigurations. Reconfigurations B and E showed a higher offset ($>$25~mm), whereas reconfiguration D showed the smallest offset of only 11.61~mm. The offset of reconfiguration E can be explained by the significantly larger $\Delta_z$ compared to $\Delta_x$ and $\Delta_y$, where the reconfiguration was at a large distance from the origin, resulting in a large moment on the malleable link. Over the workspace exploration, this caused the malleable to slightly deflect and translate the workspace in the $z$ axis, which can be seen in the plotted reconfiguration E in Fig.~\ref{expworkspace}. In comparison, reconfiguration B showed a similar offset across all axis. When compared as a percentage of the maximum radius of the workspace of the malleable robot (1100~mm), we can compute the percentage error of the normal offset for each reconfiguration, shown in Table~\ref{results-accuracy-alignment}. The results obtained were fairly consistent, as across all reconfigurations the error was 1.8\%, with the largest error shown by reconfiguration E (2.5\%) and the smallest error shown by configuration D (1.1\%).

While intrinsic malleable robots are an unexplored area, we can instead compare this accuracy to existing continuum robots of similar structure, which are typically actuated and significantly smaller than malleable robots. Using pneumatics or tendons an accuracy of $<$15~mm can be achieved \cite{kang2016design,qin2021snake}, and using concentric tubes an accuracy of $<$5~mm can be achieved \cite{su2012mri,su2016concentric}. We see the alignment accuracies achieved manually with the malleable robot are close to the accuracy expected with pnuematic and tendon driven robots, which is impressive considering the significantly larger size (and therefore weight) of the malleable robot. The manual alignment process however does raise questions regarding the resulting robot accuracy, where significant improvements must be made for use in high accuracy environments. One method explored for improving the alignment accuracy of extrinsic malleable robots was the use of an Augmented Reality (AR) headset to guide the user in the reconfiguration of the robot \cite{ranne2021augmented}. With the addition of the AR headset it was possible to reduce the normal error to $<$10~mm, roughly half of what was achieved manually without the headset.

Looking at the shapes of the produced workspaces in Fig.~\ref{expworkspace}, we observe most of the reconfigurations produced an experimental workspace matching the theoretical, with reconfigurations A, B, and C generating an almost identical workspace. Reconfiguration E shows the workspace translated in the $z$ axis, due to aforementioned reasons, while reconfiguration D shows the largest variation compared to the theoretical workspace, most likely due to inconsistent interpoint distance offsets in the reconfiguration. This suggests some workspaces are more susceptible to inaccuracies for the end effector, as the smallest end effector offset was shown for this reconfiguration. One explanation for this is the size of the workspace, where it is significantly smaller than the other reconfigurations. Reconfiguration A also has a small workspace, and also demonstrated a smaller (1.5\%) percentage error for the end effector offset compared to the other reconfigurations.

\begin{figure}[t!]
    \centering
    \includegraphics[width=\columnwidth]{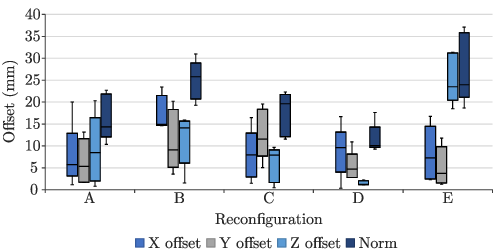}
    \caption{Boxplots of the raw alignment accuracy results for $\Delta_x$, $\Delta_y$, $\Delta_z$, and Normal.}
    \label{boxplot}
\end{figure}

The distribution of the axes offsets and normal offset of all of the repeats for the alignment accuracy experiment are shown in the form of box plots in Fig.~\ref{boxplot}. Here we see a similar distribution of offsets (IQR = 5-10~mm) across all reconfigurations, except for reconfiguration D which shows a much smaller distribution for the $\Delta_y$, $\Delta_z$, and resulting norm. We can also confirm the reasoning behind the increased norm of reconfiguration E is due to the significantly higher offset of $\Delta_z$.

To conclude, the payload performance of the malleable robot was assessed. From the results in Fig.~\ref{payload}, we observe the PUMA robot type of topology performed the best, due to the distal joint of the robot being located directly above the base joint. In comparison, the Spherical and General articulated performed similarly with the distal joint located further from the base, while the SCARA type performed the worst with on average an additional 50~mm of deflection over the PUMA type, with the distal joint located the furthest from the base. While in an ideal case a serial robot would show a significantly smaller deflection, the soft variable stiffness nature of the robot, combined with the scale of the serial robot (maximum working diameter 2.2~m), results in the expected continuous deformation with increased load seen. Thus, when selecting a reconfiguration to achieve based on the topology generation method described, it is recommended to select a reconfiguration where the distal joint is located as close to directly above the base joint for improved accuracy and payload performance.

\section{Conclusion}
In this paper we presented the design of a 2-DOF malleable robot, detailing innovations in layer jamming termination and joint design. We presented the generation of the forward and inverse kinematics of a 2-DOF malleable robot using distance geometry. We also presented the ability to define an end-effector vector in space and generate the relevant geometric robot topology, defined by 4 interpoint distances that define the reconfiguration of the malleable link, required to achieve said end effector location. The proposed kinematics and topology reconfiguration were evaluated experimentally, along with the manual alignment learning and payload capability of the malleable robot. Considerations were identified for reconfiguring the robot, namely the joint angle limits and distal joint position, and how they significantly (negatively) affect the ability to achieve the desired end effector location and payload. In reconfiguring the robot, it was observed that 5 repeated reconfigurations were necessary to train a user in obtaining an accurate alignment, with overall accuracy improving with experience. From the computed optimal topologies, accurate alignments ($<$5~mm interpoint distance offsets) were achieved for all reconfigurations, and quite accurate end effector positions ($<$11~mm axes offset) were achieved. 


\bibliographystyle{IEEEtran}
\bibliography{references}

%

\newpage
\begin{IEEEbiography}[{\includegraphics[width=1in,height=1.25in,clip,keepaspectratio]{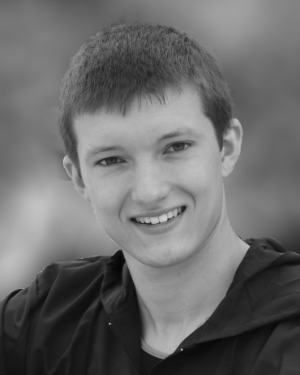}}]{Angus B. Clark} (S'17) received the M.Eng. degree in mechanical engineering from the University of Southampton, Southampton, UK, and the Ph.D. degree in design engineering research specialised in robotics at Imperial College London, London, UK. His research interest includes the development of variable stiffness malleable robots, surgical robotics, manipulation and grasping, and exoskeletons and prosthetics.
\end{IEEEbiography}

\begin{IEEEbiography}[{\includegraphics[width=1in,height=1.25in,clip,keepaspectratio]{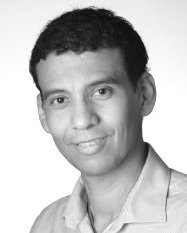}}]{Nicolas Rojas} (M’13) received the B.Sc. degree (Hons.) in electronics engineering from Javeriana University, Cali, Colombia; the M.Sc. degree in industrial engineering from University of Los Andes, Bogota, Colombia; and the Ph.D. degree (Hons.) in robotics from Polytechnic University of Catalonia, Barcelona, Spain. He was a Post-Doctoral Research Fellow with the SUTD–MIT International Design Center, Singapore; a Post-Doctoral Associate with the Department of Mechanical Engineering and Materials Science, Yale University, New Haven, CT, USA; and a Lecturer in mechatronics with the Department of Engineering and Design, University of Sussex, Brighton, UK. He has been a Lecturer (Assistant Professor) with the Dyson School of Design Engineering, Imperial College London, London, UK, where he has led the Robotic manipulation: Engineering, Design, and Science Laboratory (REDS Lab) since 2017. His research interests include robotic manipulation, dexterous manipulation, soft robots, and grasping.
\end{IEEEbiography}



*
\enlargethispage{-5in}

\end{document}